\documentclass[11pt,a4paper]{article}
\usepackage[table]{xcolor} 
\usepackage{times,latexsym}
\usepackage{url}
\usepackage[T2A,T1]{fontenc} 

\usepackage[acceptedWithA]{tacl2021v1}

\usepackage{xspace,mfirstuc,tabulary}
\usepackage{booktabs}
\usepackage[utf8]{inputenc}   
\usepackage[russian,english]{babel}  
\usepackage{adjustbox}
\usepackage{amsmath, amsfonts}
\usepackage{algorithm}
\usepackage{array}
\usepackage{tabularx}
\usepackage[noend]{algpseudocode}
\usepackage{float} 
\usepackage{wrapfig} 
\usepackage{arydshln} 
\usepackage{tcolorbox}
\usepackage{caption}
\usepackage{subcaption}
\usepackage{afterpage}
\usepackage{multirow}

\newcommand{\mv}[1]{\textcolor{black}{#1}}
\newcommand{\mvv}[1]{\textcolor{black}{#1}}

\newif\iftaclinstructions
\taclinstructionsfalse 
\iftaclinstructions
\renewcommand{\confidential}{}
\renewcommand{\anonsubtext}{(No author info supplied here, for consistency with
TACL-submission anonymization requirements)}
\newcommand{\instr}
\fi

\iftaclpubformat 

\else

\fi

\title{Navigating Cultural Chasms: Exploring and Unlocking the Cultural POV of Text-To-Image Models}

\author{
 Mor Ventura\textsuperscript{1} 
 \hspace{0.5cm} Eyal Ben-David\textsuperscript{1} 
 \hspace{0.5cm} Anna Korhonen\textsuperscript{2} 
 \hspace{0.5cm} Roi Reichart\textsuperscript{1} \\
 \textsuperscript{1}Faculty of Data and Decision Sciences, Technion, IIT \\
 \textsuperscript{2}Language Technology Lab, University of Cambridge, UK \\
  \texttt{mor.ventura@campus.technion.ac.il} 
}

\date{}

\begin{document}
\selectlanguage{english}
\maketitle

\begin{abstract}

Text-To-Image (TTI) models, such as DALL-E and StableDiffusion, have demonstrated remarkable prompt-based image generation capabilities. Multilingual encoders may have a substantial impact on the cultural agency of these models, as language is a conduit of culture. In this study, we explore the cultural perception embedded in TTI models by characterizing culture across three tiers: cultural dimensions, cultural domains, and cultural concepts. Based on this ontology, we derive prompt templates to unlock the cultural knowledge in TTI models, and propose a comprehensive suite of evaluation techniques, including intrinsic evaluations using the CLIP space, extrinsic evaluations with a Visual-Question-Answer (VQA) models and human assessments, to evaluate the cultural content of TTI-generated images. To bolster our research, we introduce the CulText2I dataset, based on \mv{six} diverse TTI models and spanning ten languages. Our experiments provide insights regarding \textit{Do}, \textit{What}, \textit{Which} and \textit{How} research questions about the nature of cultural encoding in TTI models, paving the way for cross-cultural applications of these models. 
\footnote{Our code and data are available at \url{https://github.com/venturamor/CulText-2-I}.}
\end{abstract}


\section{Introduction}
\label{sec:intro}
\textit{``We seldom realize, for example that our most private thoughts and emotions are not actually our own. For we think in terms of languages and images which we did not invent, but which were given to us by our society.''} \citep{watts1989taboo} \\

\begin{minipage}{0.45\textwidth} 
\begin{figure}[H] 
    \centering
    \includegraphics[width=\linewidth]{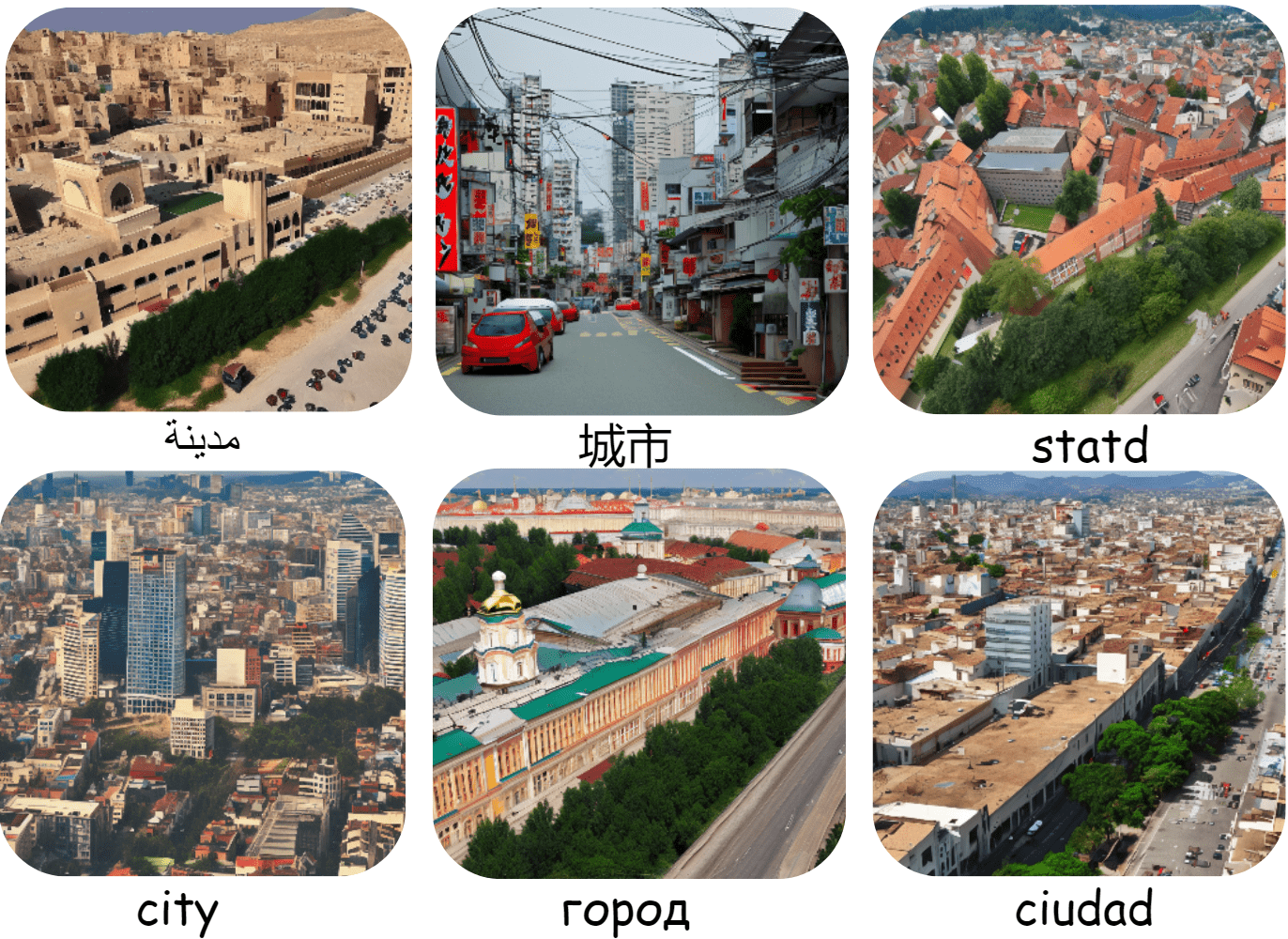}
    \vspace{-15pt}
    \caption{StableDiffusion 2.1v images of \textit{“A photo of <city>”}, while \textit{city} is translated to (left to right) Arabic, Chinese, German (top) English, Russian, Spanish (bottom).}
    \label{fig:city_diffrences}
\end{figure}
\end{minipage}
\\

Generative Text-To-Image models (TTI, e.g., DALL-E \citep{ramesh2021zero_delle, ramesh2022hierarchical_dalle2} and StableDiffusion \citep{rombach2021highresolution_sd}) have recently witnessed a surge in popularity, due to their remarkable zero-shot capabilities. They are guided by textual prompts to generate images, offering a visual representation of their textual interpretation.

TTI models exhibit multilingual proficiency, acquired \textit{explicitly}, through the model architecture and objective,  or \textit{implicitly}, through exposure to multiple languages only (\S \ref{sec:related}). These models find widespread use in domains such as art, education, and communication, exerting substantial societal influence \citep{ko2023large, vartiainen2023using, maharana2022storydall}. Their cultural significance stems from their multilingual competence and extensive adoption, as language is a vessel for cultural identity and heritage. Indeed, \citet{yiu2023imitation} have demonstrated  AI models' pivotal role in enhancing cultural transmission.

\mv{This study aims} to gain insight into the cultural perception inherent in TTI models. We embark on a novel characterization, dissecting the complex \mv{ties} between language, culture, and TTI models. Our approach \mv{is inspired by} well-established cultural research \citep{hofstede1983dimensions, rokeach1967rokeach, WVS_Survey}, allowing us to systematically deconstruct the wide notion of culture across three tiers, progressing from the broader to the finer levels of abstractness: \textit{cultural dimensions, cultural domains, and cultural concepts}.

This ontology allows us to derive prompt templates, with which we aim to unlock the cultural knowledge encoded within TTI models, and a suit of evaluation measures which reflects different aspects of the cultural information in a generated image. These methods consist of intrinsic evaluations using the CLIP \mv{\cite{clip_radford2021learning, openclip_cherti2022reproducible}} space, extrinsic evaluations with VQA models \mv{\cite{li2023blip, achiam2023gpt}}, and human assessments. 

\mv{To address the lack of an appropriate dataset, we introduce CulText2I, comprising images generated by six distinct TTI models, varying in multilingual capability and architecture (\S\ref{sec:related}). These models include StableDiffusion 2.1v and 1.4v, AltDiffusion, DeepFloyd, DALL-E, and a Llama 2 + SD 1.4 UNet model based on Llavi-Bridge \citep{rombach2021highresolution_sd, ye2023altdiffusion, deepfloydlink, ramesh2022hierarchical_dalle2, zhao2024bridging}. We generate images using prompts representing identical cultural concepts across ten languages.}

In \S \ref{sec:rqs} we present our research questions, aiming to address the way cultural knowledge is encoded by and can be unlocked in TTI models, the effective ways of unlocking this knowledge and the resulting conclusions about the world cultures. \S \ref{sec:culture_cont} presents our cultural ontology, and then \S \ref{sec:culture} and \S \ref{sec:evaluation} present the derived prompt templates and evaluation measures, respectively. \S \ref{sec:exp_setup}, \S \ref{sec:results} \mvv{and \S \ref{sec:ablation_analysis} present our experiments, results and potential factors behind our key findings}, which demonstrate the cultural capacity of TTI models, the important role of the multilingual textual encoder, and the impact of the different unlocking decisions as manifested by the prompt design.
By interrogating the cultural nuances within TTI models, we hope to challenge existing NLP paradigms and inspire innovative applications that harness the models' potential for cross-cultural understanding.




\section{Related Work}
\label{sec:related}

\paragraph{Generative Text-To-Image Models}
TTI models typically incorporate two core components: an \textit{LM-based text encoder}, which interprets and processes linguistic inputs; and an \textit{image generator} (typically based on diffusion) that synthesizes corresponding images.

Multilingual text encoders \citep{mbert, xue2020mt5, xlmr_goyal2021larger, scao2022bloom} opened the door to a wide range of cultural influences on TTI models, which are at the heart of this study. The multilingual capabilities, \mv{varying in their semantic interpretation and how well they align images to the requested concept in the prompt (i.e. \textit{conceptual coverage}) \cite{saxon-wang-2023-multilingual}} are acquired either by \textit{explicit} training objectives (e.g., DeepFloyd IF \citep{deepfloydlink} and AltDiffusion \citep{ye2023altdiffusion}), such as in the training of XLM-R and T5, or  \textit{implicitly} -- only through exposure to different languages in their training corpus, as in the case of StableDiffusion v2.1 \mv{or v1.4} \citep{rombach2021highresolution_sd} which employ CLIP text encoder.\footnote {For some models (e.g., DALL-E \citep{ramesh2022hierarchical_dalle2})) the nature of the multilingual encoder is unknown.} 

\paragraph{Foundational Frameworks in Cultural Studies}
\citet{rokeach1967rokeach} introduced the Rokeach Value Survey (RVS), illuminating how classified values shape behaviors at both individual and societal levels. Building on this, \citet{hofstede1983dimensions} laid grounding work in studying cultural variances by introducing key cultural dimensions like femininity versus masculinity, fostering a systematic approach to exploring cultural differences. \citet{bond1988finding} identified 36 ``universal values'', such as love and freedom, underscoring the shared human values across diverse cultures and regions. \citet{schwartz1994beyond} further refined our understanding by proposing a universal framework of ten fundamental human values, revealing how they are prioritized and interpreted diversely across cultures, impacting behaviors and attitudes.
Later advancements by \citet{triandis1998converging} and \citet{mccrae2002five} further enriched the domain of cross-cultural psychology. Triandis emphasized the nuances of individualism and collectivism, while McCrae delved into the variances in the manifestations of the Big Five personality traits across different cultural settings. Complementing these, the World Values Survey (WSV, \citet{WVS_Survey}) introduced an innovative cultural map, portraying the global shift towards more secular and self-expression values as societies advance and prosper.

This work synthesizes the cultural literature presented above to establish a comprehensive repository of cultural concepts and dimensions.
In \S\ref{sec:culture_cont}, we introduce our culture characterization \mv{for detailed} analyses of the representation and perception of cultural perspectives within TTI models. 

\paragraph{Culture in LMs and TTI models}
With the burgeoning interest in Pre-trained Language Models (PLMs) and TTI diffusion models, there is increasing scrutiny of the cultural gaps and biases inherent within these models. These gaps manifest as discrepancies in the representation of norms, values, beliefs, and practices across diverse cultures \citep{Rao2024NORMADAB, Cultural_Incongruencies, struppek2022biased, DBLP:journals/natmi/AbidFZ21,  DBLP:conf/emnlp/AhnO21, touileb2022occupational, smith-etal-2022-im}. 
\citet{arora2022probing} and \citet{ramezani2023knowledge} explored the cross-cultural values and moral norms in PLMs and assessed their alignment with theoretical frameworks, revealing a conspicuous inclination towards western norms. 
Similarly, other studies demonstrated the challenges with English probes and monolingual LMs, which diminish the representation of non-western (e.g., Arab) norms in model responses \citep{cao2023assessing, naous2023having, masoud2023cultural, atari2023humans, Putri2024CanLG}. While these works focused on the cultural implications of PLMs, detecting or mitigating cultural biases, we develop a methodology that inspects the TTI models' cultural values, and seek to understand their internal representations of cultures.

While cultural exploration in TTI model research is relatively limited, there have been advancements to enhance cultural diversity within these models. Multilingual benchmarks, focusing on Chinese and Western European/American cultures, have been introduced  \citep{liu2023cultural, liu2021visually}. Efforts to uncover cultural biases include evaluations of nationality-based stereotypes \citep{Jha2024ViSAGeAG}, skin tone biases \citep{cho2022dalleval}, and associated risks \citep{bird2023typology}. Analyses also address social biases in English-only TTI models \citep{naik2023social}, covering gender, race, age, and geography. \citet{Kannen2024BeyondAC} and \citet{basu2023inspecting} have evaluated the cultural competence of these models. Our work extends beyond Western tendencies, focusing on multilingual TTI models and exploring fine-grained cultural concepts and dimensions.

\section{Research Questions and Overview}
\label{sec:rqs}

Our primary research inquiry revolves around:  
\textbf{How does multilingual TTI models capture cultural differences?}
To delve into this overarching question, we craft four research questions:

\begin{itemize}
    \item \textit{\underline{RQ1}: \textbf{Do} TTI models encode cultural knowledge?}
    \item \textit{\underline{RQ2}: \textbf{What} are the cultural dimensions encoded in TTI models?}
    \item \textit{\underline{RQ3}: \textbf{Which} cultures are more similar according to the model?}
    \item \textit{\underline{RQ4}: \textbf{How} to unlock the cultural knowledge?}
\end{itemize}

RQs 1-3 form a hierarchy, with each question building upon the previous one. RQ4 stands as an independent, high-level question. Our methodology consists of three pillars: (1) crafting a cultural ontology; (2) experimenting with TTI models featuring diverse multilingual text encoders, and (3) employing a triad of evaluation methodologies: \textit{intrinsic evaluation} using OpenClip, \textit{extrinsic evaluation} involving Visual Question Answering (VQA) models, and \textit{human assessment}.

\section{Cultural Ontology}
\label{sec:culture_cont}

\begin{table*}[htbp]
  \centering
  \caption{Prompt Templates: Language and Gibberish prompts}
  \label{tab:prompt_templates} 
    \begin{adjustbox}{width=\textwidth}  
      \begin{tabular}{l|l |l}
        \toprule
        Prompt Description & Prompt Template (T - translated, EN - English) & Example \\
        \midrule
        English Reference & EN : ``a photo of <concept>'' &  ``a photo of food'' \\
        \midrule
        Fully Translated Prompt & T : ``a photo of <concept>'' & \begin{otherlanguage}{russian} “фото еда” \end{otherlanguage} \\
        Translated Concept & EN : ``a photo of'' + T : <concept> & ``a photo of \begin{otherlanguage}{russian} еда\end{otherlanguage}'' \\
        English with Nation & EN : ``a photo of <nationality> <concept>'' & ``a photo of Russian food'' \\
        \hdashline
    English with Gibberish & EN : ``a photo of <concept>'' + T : ``<gibberish>'' & ``a photo of food \begin{otherlanguage}{russian} йкуаскымдо\end{otherlanguage}'' \\
    \bottomrule
  \end{tabular}
  \end{adjustbox}
\end{table*}

\begin{figure*}[!htb]
  \centering
  \includegraphics[width=1.0\textwidth]{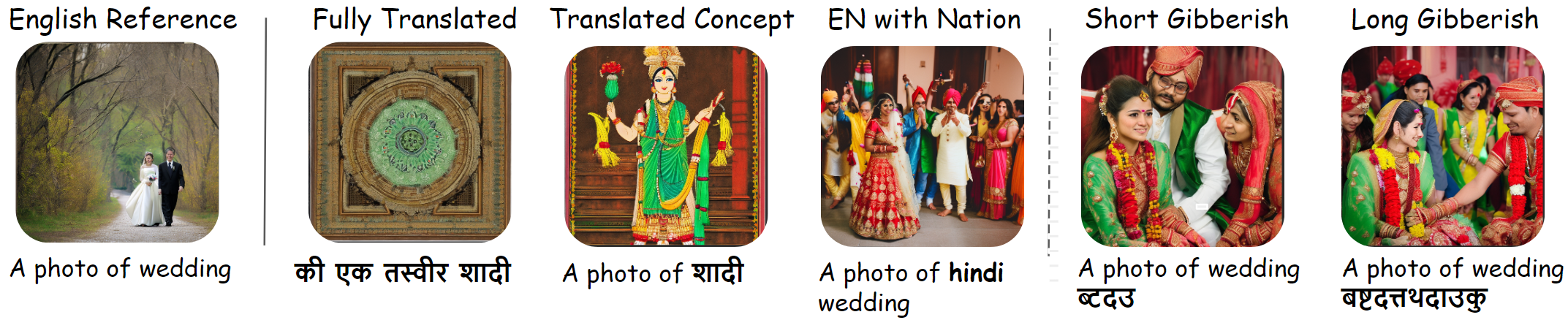}
  \vspace{-25pt}
  \caption{StableDiffusion images generated from all the prompt templates (PTs) for the cultural concept (CC) of \textit{Wedding} and the Hindi language.} 
  \label{fig:indian_wedding_prompt}
\end{figure*}


We aim to design an ontology that will allow us to (1) consolidate diverse perspectives on culture; and (2) quantitatively assess cultural aspects within the context of TTI models. Research on cultural definitions is typically based on breaking down the big idea of culture into different aspects like individualism or science, which often involve intricate and abstract details or queries unsuitable for visual examination. 
To utilize culture studies to our needs, we hence develop two key pillars: (a) \textbf{\textit{cultural domains}}, comprising  \textbf{\textit{cultural concepts}}; and (b) \textbf{\textit{cultural dimensions}}.

\paragraph{Cultural Domains and Concepts.} Drawing inspiration from established categorizations in works like \citet{hofstede1983dimensions, rokeach1967rokeach, WVS_Survey}, we \mv{combine} ten common and \mv{broad} aspects to form the cultural domains. Each domain reflects a collection of values, tendencies, and beliefs, which we represent through concise concepts. For instance, the cultural concept of \textit{Heaven} in the \textit{Religion} domain is derived from the question in the religion section of the World Values Survey, which asks: ``Do you believe in heaven?''. We define twelve domains and 200 cultural concepts (see Table \ref{table:cc_all} in the Appendix).
These domains include \textit{Moral Discipline and Social Values} (example concepts: Housewife, Divorce), \textit{Education} (Teacher, Engineer), \textit{Economy} (Market, Job), \textit{Religion} (God, Wedding), \textit{Health} (Doctor, Medicine), \textit{Security} (War, Weapon), \textit{Aesthetics} (Art, Fashion), \textit{Material Culture} (Car, Camera), \textit{Personality Characteristics and Emotions} (Lazy person, Proud person), and \textit{Social Capital and Organizational Membership} (City, Police).




\paragraph{Cultural Dimensions.} 

Certain cultural aspects function more like axes (e.g. from Individualism to Collectivism) than as comprehensive domains (e.g. Science). We grouped these aspects under the category of cultural dimensions. \mv{The dimensions we use are defined as follows: (1) \textit{Traditional versus Rational} values;\footnote{The original term is \textit{Secular}-\textit{Rational}.} (2) \textit{Survival versus Self-expression} values \citep{WVS_Survey}; 
(3) \textit{Critical versus Kindness};\footnote{\citet{mccrae2002five} originally named this dimension by \textit{Neuroticism versus Adjustment}.} (4) \textit{Extraversion versus Introversion} \citep{mccrae2002five}; (5) \textit{Modern versus Ancient} values; (6) \textit{Masculine versus Feminine} attributes; (7) \textit{Individualism versus Collectivism} and (8) \textit{Nature versus Human} \citep{hofstede1983dimensions, schwartz1994beyond}.}\footnote{\citet{schwartz1994beyond} originally included this in the values \textit{Universalism} and \textit{Harmony}.}
\mv{The cultural dimensions don't cover all suggested \mvv{aspects} from the original research. We focused on aspects that are more visually representable and quantifiable.
}
\section{Unlocking Culture in TTI models}
\label{sec:culture}

\begin{figure*} 
  \centering
  \includegraphics[width=0.95\textwidth]{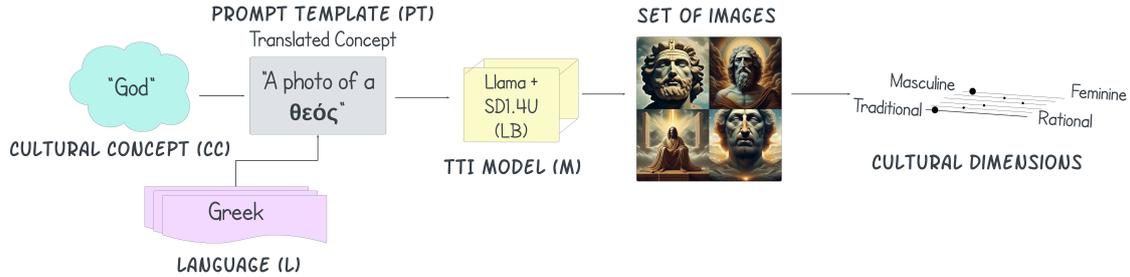}
  \caption{TTI model workflow scheme. The visual representations of each Cultural Concept (CC) are image sets generated with different languages (L) and prompt templates (PTs) by different TTI models (M). \mvv{Then, the images' cultural content is evaluated.} Here, for example, CC is \textit{God}, PT is \textit{Translated Concept}, \mv{M is Llama2 + SD1.4 UNet (LB) and the \mvv{evaluation uses} the cultural dimensions metrics \mvv{(\S \ref{sec:auto_metrics})}}.}
  \label{fig:drawio}
\end{figure*}

We now introduce our prompt templates which feed the cultural concepts as input into the TTI models, aiming to unlock the effect of different cultures on their outputs.


\paragraph{Cultural Concepts and Dimensions.}

In \S \ref{sec:culture_cont} we defined cultural concepts, denoted below with  $\mathrm{\left\{ \textit{CC} \right\}}_{i=1}^{200}$, and cultural dimensions, denoted as $\mathrm{\left\{ \textit{CDM} \right\}}_{i=1}^{8}$. Cultural concepts are dynamic parts of the TTI model templated input  (see Table~\ref{tab:prompt_templates}). 
Every \textit{CC} is expressed by one or two English words (e.g., \textit{Food}), acting as a \mv{concise} representation of a more expansive domain (e.g., \textit{Aesthetics}). In contrast, the cultural dimensions are used in our outcome measures but not in the prompts.


\paragraph{Prompt Templates.}
We construct five prompt templates (PTs; see Table \ref{tab:prompt_templates} and resulting images in Figure \ref{fig:indian_wedding_prompt}).
These templates aim to discern the cultural implications carried solely by the linguistic characters of the language in question. In our setup, a $\textit{PT}$ is defined as a function $\phi$ of the target language, $\textit{L}$, and a cultural concept, $\textit{CC}$: $\textit{PT} = \phi(\textit{L}, \textit{CC})$. The first two PTs (\textit{Translated PTs}) with the third (all together - \textit{Language PTs}) enable us to investigate whether the language can convey cultural information, while the last PT (\textit{Gibberish PT}) aims to explore if linguistic characters alone can convey such information, especially when the language lacks extensive translated data in the TTI models' training data. \mv{Interestingly, in our experiments we found the more non-English information (words, characters) the prompt contains, the lower the conceptual coverage (see Figure \ref{fig:pt_conceptual_coverage} \mvv{and details in Appendices \ref{subsec:coverage_formula} and \ref{app:cc_tangible})}.}

\paragraph{TTI Model Workflow.}
A TTI model, $\textit{M}$, operates by receiving a textual prompt as input, denoted as $\textit{In}$, and in turn, generating a corresponding set of images as output. We form $\textit{In}$ through prompt templates, $\textit{PTs}$ (Table \ref{tab:prompt_templates}), to study how changing parts of $\textit{In}$ affects the image generated by the model ($\textit{M}$). By employing a $\textit{PT}$, we are able to keep all elements of the input constant except the one under examination. As depicted in Figure~\ref{fig:drawio}, the $\textit{PT}$ shapes the input ($\textit{In}$) to $\textit{M}$, culminating in a generated image that reflects the interplay between cultural concepts within the model's parameters \mvv{(See examples of generated images in Figures \ref{dalle_deepf_food_family_music}, \ref{wedding_sd_ad} in the Appendix).}

\section{Evaluating Cultural Aspects in Images}
\label{sec:evaluation}
\mvv{In this section, we discuss how we evaluate the images' cultural content.} We employ two \textbf{automatic} measures for cultural characteristics (\textbf{intrinsic} and \textbf{extrinsic}), as well as \textbf{human assessment}.

\subsection{Automatic Metrics}
\label{sec:auto_metrics}

\begin{table*} [htbp]
\begin{adjustbox}{max width=1.0\textwidth} 
\centering
\setlength{\tabcolsep}{2pt} 
\renewcommand{\arraystretch}{1.2} 
\begin{tabular}{|l|l|l|c|}
\hline
\textbf{Metric} & \textbf{$X$: Template } & \textbf{$X$: Instance} & \textbf{$D$} \\
\hline
\rowcolor{gray!30} 
\multicolumn{4}{|c|}{\textbf{Identifying Cultural Origin [RQ1, RQ4]}} \\ 
\hline
National Association (NA) & \textit{a photo with <national> style} & \textit{a photo with spanish style}& $\text{softmax}\left(\Vec{\text{D}}\right)$ \\
Extrinsic NA (XNA) & \textit{What is the country of origin for the} & - & VQA, Majority \\
& \textit{depicted photo?} & &   \\
\hline
\rowcolor{gray!30} 
\multicolumn{4}{|c|}{\textbf{Depicting Cultural Dimensions [RQ2]}} \\ 
\hline
 Cultural Dimensions Projection (DP) & \textit{a photo with <cultural dimension> aspects} & \textit{a photo with modernity aspects} & $\left(\text{I} \cdot X^t\right)$ \\
 Extrinsic DP (XDP ($d_0$, $d_1$)) &  \textit{Are there more <$d_0$> features in the photo}  &  \textit{Are there more modern features in the photo} & VQA, Majority \\
& \textit{or more <$d_1$>?} & \textit{or more ancient?} &  \\
\hline
\rowcolor{gray!30} 
\multicolumn{4}{|c|}{\textbf{Finding Cultural Similarities [RQ3]}} \\ 
\hline
Cultural Distance (CD) & $EN$\textit{:a photo of <cultural concept>} & \textit{a photo of city} & $1 - \left(\text{I} \cdot X^t\right)$  \\
Cross-Cultural Similarity (CCS ($l1$, $l2$)) & \textit{a photo of} $T_{l2}$\textit{:<cultural concept>} & \textit{a photo of ciudad} &  $I_{l1} \cdot (X^v)_{l2}$ \\
\hline


\end{tabular}
\end{adjustbox}
\vspace{-5pt}
\caption{Automatic Metrics - Grouped by their aims and research questions. $I$ corresponds to the OpenClip visual representation of an image in the set generated by a TTI model. $X^t$ and $X^v$ stand for the textual and visual representations of the baseline prompt ($X$:Instance), respectively. $l1$ encodes the language of the inspected $I$ while $l2$ encodes the other language of the evaluation prompt $X$. $\Vec{\text{D}} = \left[\textit{I} \cdot X^t_1, \textit{I} \cdot X^t_2,...,\textit{I} \cdot X^t_k\right]$, for the $k$ nationalities. \mv{Finally, $T$ stands for 'Translated', and $d_0$ and $d_1$ stand for the dimension extremes, respectively}.
} 
\label{table:automatic_metrics}
\end{table*}

We introduce \mv{six } metrics, corresponding to our research questions (Table~\ref{table:automatic_metrics} and \S \ref{sec:rqs}), which fall into two categories: \textit{intrinsic}, utilizing internal representations, and \textit{extrinsic}, relying on an external Visual Question Answering (VQA) model. Consistent with previous culture-in-TTI research \citep{wang2023exploring, naik2023social, liu2023cultural}, we construct intrinsic measures using both textual and visual representations from image-text encoders (OpenClip in our case).
The intrinsic metrics follow the equation: $\mathbb{D} \left(\textit{I} \cdot \textit{X}\right)$.



Here, $I$ is a visual representation of an image generated by a TTI model in response to a prompt of interest. $X$ is either a textual representation ($X^t$) or a visual representation ($X^v$) of the metric's baseline prompt (the \textbf{$X$: Instance } column of the table), and the choice of $X$ being textual or visual representation is metric specific. $D$ is the metric operator, that is applied to the cosine similarity scores between $I$ and $X$. The eventual metric is an average of the $D$ values over the $n$ images in the set generated by the TTI model in response to a given prompt (see \S \ref{sec:culture}). This section will exemplify our metrics with a running example of $I$ curated from the prompt: ``a photo of <stadt>'' (\textit{stadt} is the German word for \textit{city}). 

\mv{\underline{\textit{Identifying Cultural Origin [RQ1, RQ4]}}} The first metric, \textit{\textbf{National Association (NA)}}, aims to identify the origin culture of the image. Intrinsically (top table row), we calculate the cosine similarity scores between the generated image representation ($I$) and the textual representations of the nationality templates ($X$, e.g. a national prompt for the ``Spanish'' style). We do so with all the nationalities\footnote{\mv{see Table \ref{tab:lang_nationalities} in the Appendix.}} in our data and apply the softmax operator on the scores vector. If the image does correspond to the German culture, we would expect the German coordinate in the vector to be high. Extrinsically (\textbf{\textit{XNA}}, \mv{second} row), we direct a query to the VQA model, inquiring about the origin country of the image $I$. Subsequently, we compute the majority vote over the set of generated images. If there is no clear majority, the XNA answer is ``can’t tell''. \mv{The score is the fraction of times a question on the image is answered correctly by the majority vote.}

\mv{\underline{\textit{Depicting Cultural Dimensions [RQ2]}}}: By the \textit{\textbf{Cultural Dimensions Projection metrics (DP, XDP)}}, we evaluate the extent to which cultural dimensions are manifest within the images. As can be seen in rows \mv{3} and \mv{4} of the table, the computation is very similar to the above nationality measures (NA and XNA).

\mv{\underline{Finding Cultural Similarities [RQ3]}}: Inspired by \citet{hofstede1991cultures}'s definition of culture,\footnote{ ``Collective mental programming distinguishing one group from another''.} we introduce the \textit{\textbf{Cultural Distance (CD)}} and \textit{\textbf{Cross-Cultural Similarity (CCS)}} metrics to probe cultural distinctions. In CD, we assess how closely various cultures align with the English culture.\footnote{Using the EN reference in the CD measure aligns with prior studies which employed EN as a reference for western cultures \citep{atari2023humans}, and acknowledges the English predominance in the training data of our TTI models.} In CCS we measure image similarities to explore how different languages influence the visual representation of the same cultural concept ($cc$).


\subsection{Human Evaluation}
\label{sec:human_metrics}

We create a questionnaire (see questionnaire example and guidelines in Figures \ref{fig:human_ques_guide}, \ref{quastionaiire_im} in Appendix \ref{app:human_eval}) for human evaluation of cultural dimensions in images generated by TTI models. The questionnaire considers 4 languages (RU, ZH, ES, DE), 12 concepts, 3 prompt templates (English with Nation, Translated Prompt and English with Gibberish) and 3 models (2 with implicit multilingual encoding - SD and DL; and 1 with explicit multilingual encoding - AD), involving 15 annotators (3 per item) on the LabelStudio \cite{LabelStudio} platform.  
For each (TTI model, prompt template, concept) triplet we generated 4 images per language, for a total of 1728 images in the entire evaluation set. Each triplet is represented by 1 page in the questionnaire, consisting of four 4-image grids,\footnote{Figure \ref{fig:drawio} presents an example of a 4-image grid.} 1 grid per language. For each 4-image grid, annotators were asked to make 3 binary decisions - one for each of 3 \mv{arbitrary} dimensions (\textit{Modern versus Ancient}, \textit{Traditional versus Rational}, \textit{Critical versus Kindness} \mv{in} \S \ref{sec:culture_cont}), and specify the culture of origin from a given set of options (the 4 languages as well as USA).  We calculate the inter-annotator agreement (IAA) with the Fleiss kappa (\textit{Modern versus Ancient}: 0.54, \textit{Traditional versus Rational}: 0.39, \textit{Critical versus Kindness}: 0.41, national association: 0.4; on a $[-1,1]$ scale) and the agreement of human annotation (after taking the majority vote) with the ground-truth culture (74.4\%). Below we report the agreement of the automatic evaluation metrics with the majority vote between the annotators for each example.

\section{Experimental Setup}
\label{sec:exp_setup}

\paragraph{Languages.} We experiment with ten languages, serving as proxies of geographically diverse cultures: English (EN), Spanish (ES), German (DE), Russian (RU), French (FR), Greek (EL), Hebrew (IW), Arabic (AR), Chinese (ZH), and Hindi (HI). \mv{We consider two inclusion criteria: (1) The lingual coverage of TTI models \cite{saxon-wang-2023-multilingual}, and (2) Etymological Diversity. Balancing between both, we cover mainly the Indo-European language family.}

\begin{figure*}[!ht]
\centering
  \includegraphics[width=1.0\textwidth]
  {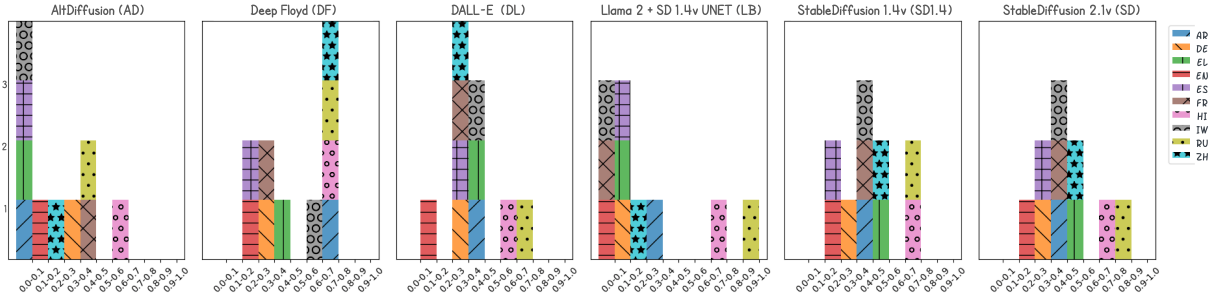}
    \vspace{-15pt}
  \caption{National Association Scores by BLIP2 (XNA) presented as \mv{histograms}. The x-axis represents bins of mean XNA scores ranging from 0 to 1 across three representative Prompt Templates (PTs): 'Translated Concept', 'EN with Nation', and 'English with Gibberish' (refer to Table~\ref{tab:prompt_templates} for details). Higher scores indicate better performance. Colors encode languages.
}
  \label{fig:blip_na_all_models}
\end{figure*}


\paragraph{Models.}


We experiment with \mv{six} SOTA TTI models, \mv{namely StableDiffusion 2.1v (SD), StableDiffusion 1.4v (SD1.4), AltDiffusion (AD), DeepFloyd (DF), DALL-E (DL) and Llama 2 + SD 1.4 UNet based on Llavi-Bridge (LB)}, differing in their multilingual textual encoders and the languages they cover (Table \ref{table:tti-languages}; \mvv{Appendix Table \ref{table:tti-tech-details}}). The multilingual capabilities of a model are affected by the languages it is trained on, and the training objective it follows. For encoders like XLM-R and T5-XXL, the multilingual aspect is \textit{explicitly} represented in the training objective, by bringing similar words in different languages closer in the learned embedding space. Also, multilingual aspects can be \textit{implicitly} represented, with different alphabets encoded differently while the objective does not impose any explicit cross-lingual constraint \mv{(e.g., as in SD)}. For the evaluation (\S\ref{sec:auto_metrics}) which requires a VQA model, we employ BLIP2 \citep{li2023blip}, which applies the Flan-T5-XL encoder.

\paragraph{Experimental Dataset.}
\label{par:dataset}

We employ the \mv{6} models to generate an image set, where each image is characterized by 4 properties: (1) the generating TTI model (M); (2) the cultural concept (CC) of interest; (3) the applied prompt template (PT); and (4) the target culture (L), encoded through the prompt, either by its language or through the culture name it mentions. We generate a \mv{$K$}-image set, \mv{for the value of $K$=4}, for each configuration of these properties, maintaining a constant initiation seed (42) for the first image in each set. This methodology yields for each TTI model T unique cultural tuples of the form (CC, PT, L, 4 \mv{images}), where the number of tuples depends on the number of languages covered by the model, see Table~\ref{table:tti-languages} ($T_{SD} = T_{AD} = 10,500$, \mv{$T_{SD1.4} = T_{LB} = T_{DF} = 6,300$} and $T_{DL}  =2,310$). \footnote{Due to API usage constraints, the DALL-E subset was limited to half of the cultural concepts (105) and three prompt templates (``English with nation'', ``Translated concept,'' and ``English with gibberish'', see Table~\ref{tab:prompt_templates}.). \mv{SD 1.4v and LB are limited to these 3 PTs as well.}} \footnote{The images in the human evaluation set of \S \ref{sec:human_metrics} are selected from this dataset.}

\begin{table}[!ht]
\centering
\tiny
\footnotesize 
\setlength{\tabcolsep}{0.8pt} 
\begin{tabular}{|c|c|c|c|c|c|c|c|c|c|c|}
\hline
& EN & ES & DE & FR & RU & EL & AR & IW & ZH & HI \\
\hline
StableDiffusion 2.1v & v & v & v & v & v & v & v & v & v & v \\
\hline
StableDiffusion 1.4v & v & v & v & v & v & v & v & v & v & v \\
\hline
Llama2 + SD1.4 Unet & v & v & v & v & v & v & v & v & v & v \\
\hline
AltDiffusion m9 & v & v & v & v & v & v & v & v & v & v \\
\hline
DeepFloyd v1.0 & v & v & v & v & v & & & & & \\
\hline
DALL-E v2 & v & v & v & v & v & & & & v & \\
\hline
\end{tabular}
%
\tiny
\footnotesize 
\setlength{\tabcolsep}{2.1pt} 
\begin{tabular}{|c|c|c|}
\hline
& Text Encoder & \mv{Objective}  \\
\hline
StableDiffusion 2.1v & OpenCLIP (ViT-H/14) & I  \\
\hline
StableDiffusion 1.4v & CLIP (ViT-L/14) & I  \\
\hline
Llama2 + SD1.4 UNET & Llama2 7b & I \\
\hline
AltDiffusion m9 & XLM-R (in AltClip) & E \\
\hline
DeepFloyd v1.0 & T5-XXL & E  \\
\hline
DALL-E v2 & Unknown & Unknown \\
\hline
\end{tabular}
\caption{Top: Model coverage of different languages. Bottom:  Model's text encoder. The coverage is based on the existence of multilingual characters (letters) in the embedding layers of the text encoder of each model (except for DALL-E, where it is based on empirical tests). Multilingual capabilities are acquired through an \textit{explicit} (E) or \textit{implicit} (I) training objective.}
\label{table:tti-languages}
\end{table}
\section{Experiments and Results}
\label{sec:results}


\paragraph{1. TTI Models Encode Cultural Identity Information (RQ1).}

Figure \ref{fig:blip_na_all_models} illustrates the extrinsic national association (XNA) scores measured on images generated by the experimental models. It is important to note that this metric uses free text answers from the VQA, which can vary widely based on national origins, making it difficult to achieve high scores. Despite \mv{that}, \mv{2} languages HI \mv{and} RU \mv{score} consistently above \mv{0.4}, with the highest mean scores across models (\mv{0.69, 0.73}), \mv{3} languages (FR, DE, AR) \mv{score} consistently above \mv{0.3} in 5 out of 6 models and only \mv{2} languages with a mean score lower than \mv{0.3} \mv{(ES and EN)}. \mvv{See detailed results in Figure \ref{table:lang_perf_models} in the Appendix}.
\mvv{We hypothesize that low English Association scores are due to the overrepresentation of English in the training data, resulting in a lack of cultural specificity. In contrast, the more limited training data for other languages is likely more culturally specific, as it is likely to be carefully selected}. Additionally, it can be associated with the global influence of American culture in the data, which may obscure distinct cultural traits, further impacting model performance in English.
Finally, the results ascertain that all the examined models can distinguish image origins.\footnote{The automatic XNA metric agrees with the human answers to the cultural origin question in 69.6\% of the cases in the human evaluation set.}

\begin{figure}[!htb]
  \centering
  \includegraphics[width=0.9\columnwidth]{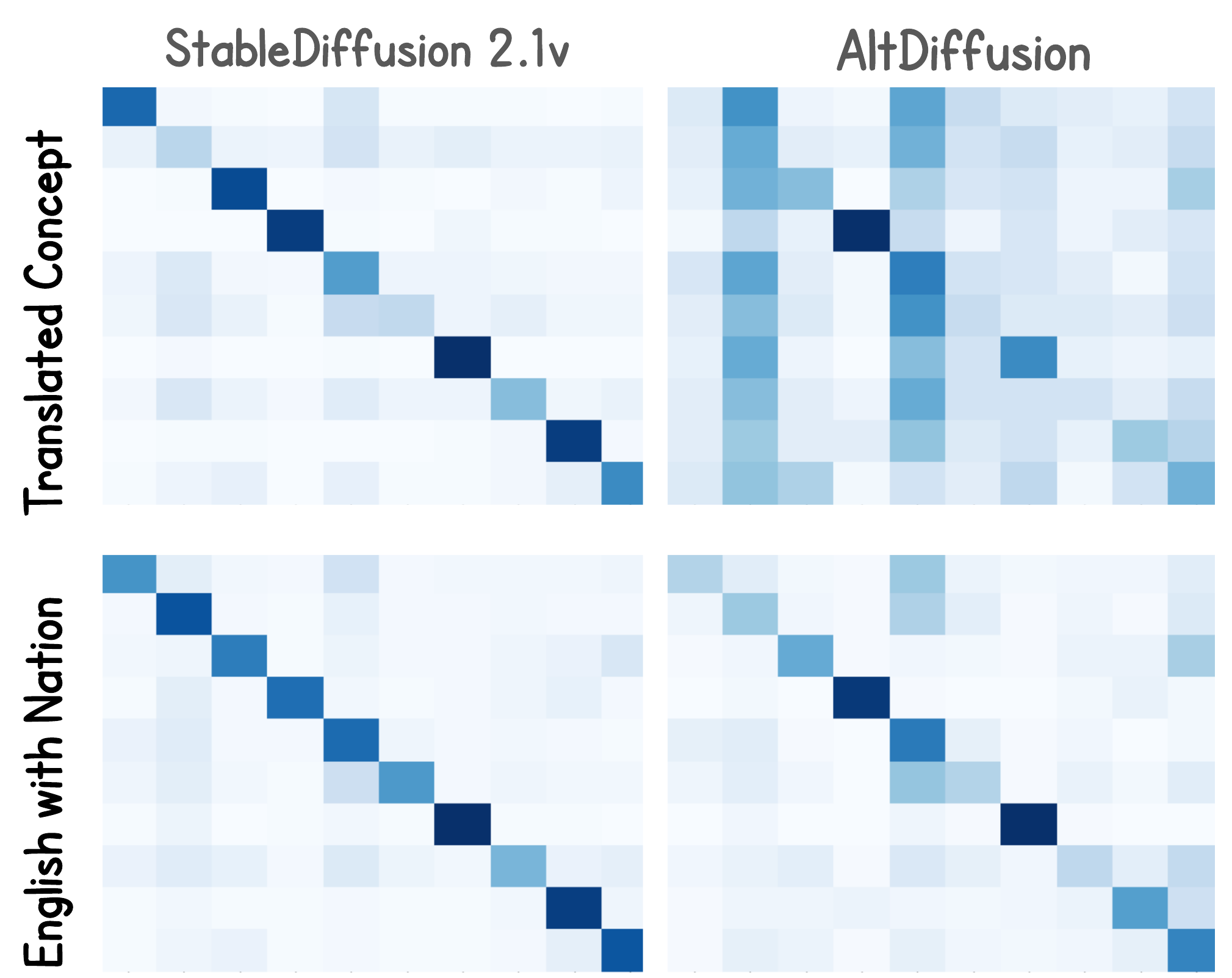}
  \caption{A \mv{confusion matrix} grid of the NA metric. Prompt Templates\footnote{The ``Fully translated PT'' is omitted after initial consistency validation.}: ``Translated Concept'' (top), ``EN with Nation'' (bottom). Models: SD (left) and AD (right). \mv{Darker} colors correspond to higher scores. y-axis: ground-truth languages. x-axis: predicted cultures. For each confusion matrix, we compute the agreement between the predicted and the ground-truth languages (Accuracy, $ACC  = \frac{1}{n}\sum_{i=1}^{n} \mathbb{I}(\text{argmax}(\text{row}_i) = i)$) to measure the cultural encoding strength of a (model, PT) pair. Languages in each grid (top-bottom, left-right): RU, EN, EL, HI, DE, FR, ZH, ES, AR, IW.
}
  \label{fig:na_openclip_nation_photo_of_sd_alt}
\end{figure}
\begin{figure*}
  \centering
  \includegraphics[width=1.05\textwidth]{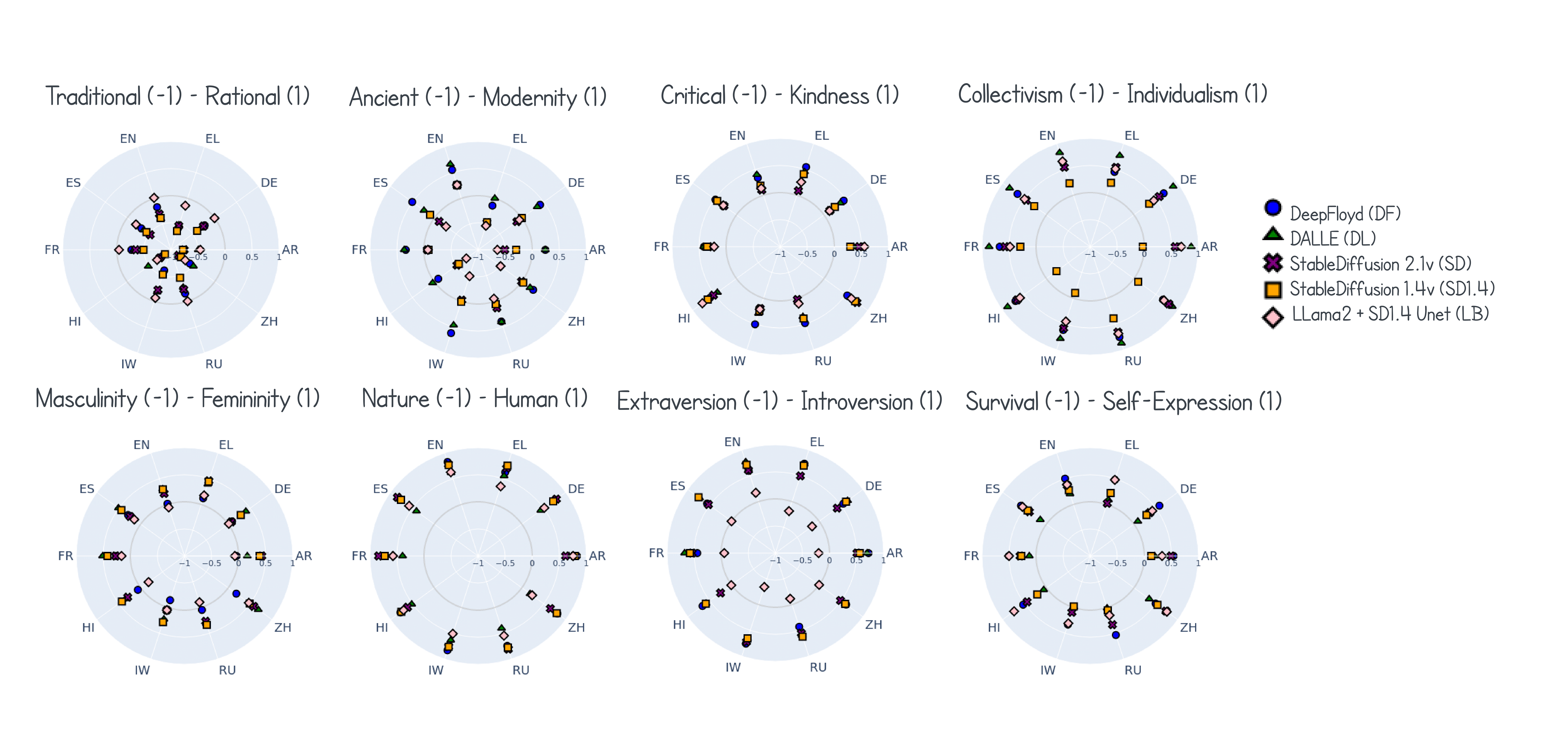}
  \vspace{-20pt}
  \caption{Radar graphs of cultural dimensions as classified by the VQA model. The scores, range [-1, 1], represent the tendency of each culture towards one of the ends of the dimension. For example, 0.6 of the Arabic images were classified as \textit{modern} and 0.3 as \textit{ancient} (0.1 as \textit{can't tell}) and hence the modernity score is 0.3.
  Circles encode cultural dimensions, markers represent models. 
  Languages appear at different angles on the perimeter. For each dimension, the negative end is at the center of the circle, while the positive end is on the perimeter. 
  For each model, results are averaged across PTs.}
  \label{fig:radar_dim_blip}
\end{figure*}

\paragraph{2. Cultural Encoding Depends On The Language-Encoding Strategy Of The Prompt And The Model (RQ4).}


Based on the intrinsic NA results presented in Figure \ref{fig:na_openclip_nation_photo_of_sd_alt}, we observe that \textbf{models with implicit multilingual encoding (SD) are better cultural encoders than models with explicit multilingual encoding (AD)}. The implied explanation is that explicit encoders bring languages closer together in the embedding space. 

Note that the main diagonal is emphasized in both the left column and the bottom row of the heatmap. This indicates that \textbf{an effective prompt (``EN with Nation'') can compensate for the effect of the encoder}. This is reflected by ACC values of 0.8 and 0.5 for explicit encoding compared to 1.0 and 1.0 for implicit encoding, for the translated prompt and the English with Nation prompt, respectively. \mvv{These patterns resonate with those observed in the other models tested.} 

Interestingly, AD with the translated prompt is biased towards the American and German cultures (100\% of the erroneously predicted cultures are classified as American or German), while the errors of AD with English with Nation are more evenly distributed.\footnote{The automatic NA metric agrees with the human answers to the cultural origin question in 75.0\% of the cases in the human evaluation set.}

\mv{Notably, since language acts as a proxy for multiple nations (e.g., English is spoken in both the US and the UK, two different cultures), we provide a second-order analysis (Figure \ref{fig:lang_proxy_nation} in the Appendix) representing the national association distribution with other nations that primarily speak these languages. This analysis \mvv{implies inherent biases within the encoding,} such as Greek images being more associated with Cyprus than Albania.}

\paragraph{3. TTI Models Encode Cultural Dimensions (RQ2).}

\begin{figure}[!htb]
  \centering
  \includegraphics[width=1.0\columnwidth]{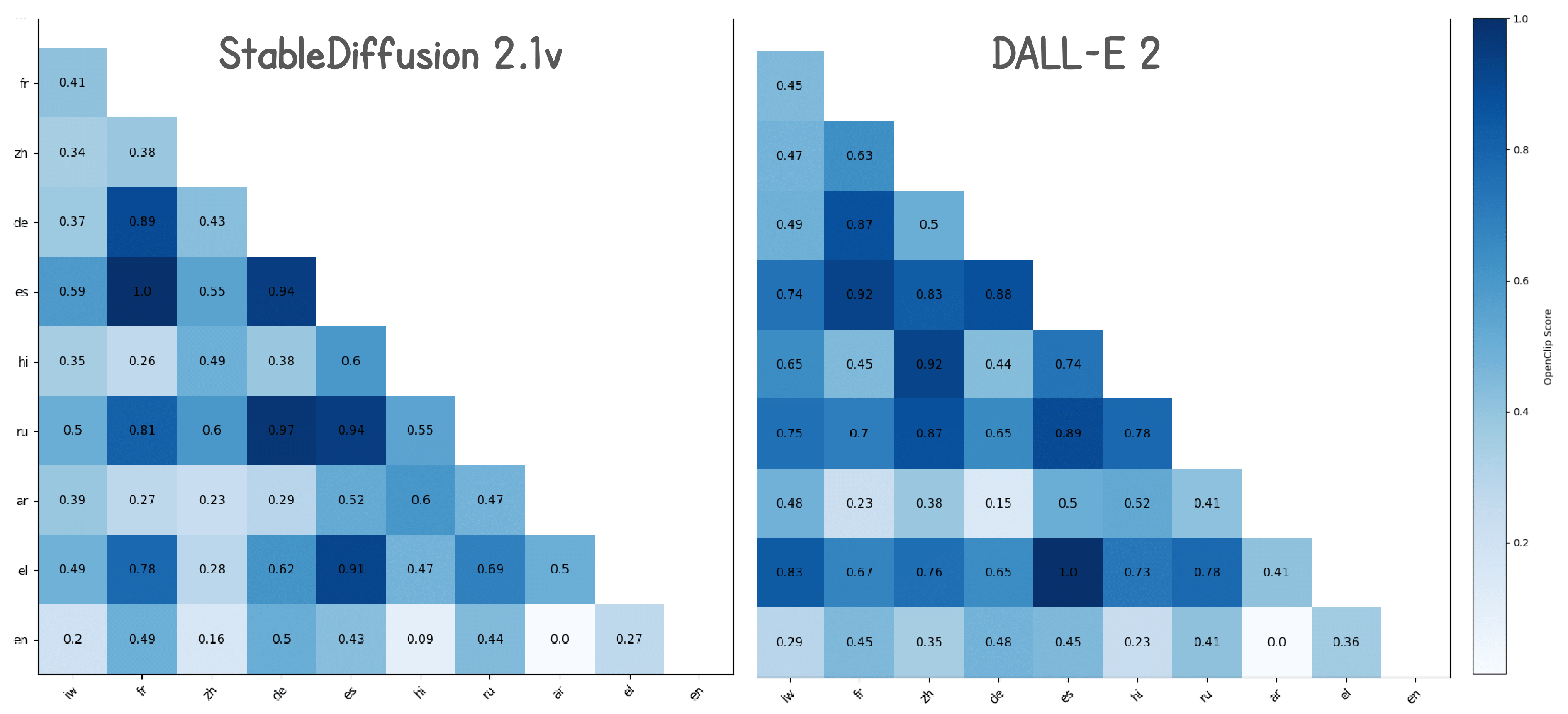}
  \vspace{-15pt}
  \caption{Cross-Cultural Similarity (CCS) analysis for the 'EN with Nation' PT. Darker values note higher similarity. \mv{The scores are normalized}.}
  \label{fig:cross_lingual_en_photo_of}
\end{figure}

\mv{\mvv{Here we show how TTI models capture cultural dimensions outlined in our ontology}. Given the challenges in defining an absolute ground-truth for cultural tendencies, we proceed with caution. We avoid direct comparisons with any such ground-truth, mindful of the potential harm such analyses could incur. Instead, in \S \ref{sec:ablation_analysis}, we carefully examine the correlations between our findings on TTI models and social science studies related to the cultures discussed.}

 \begin{figure*}[!ht]
  \centering
  \includegraphics[width=0.75\textwidth]{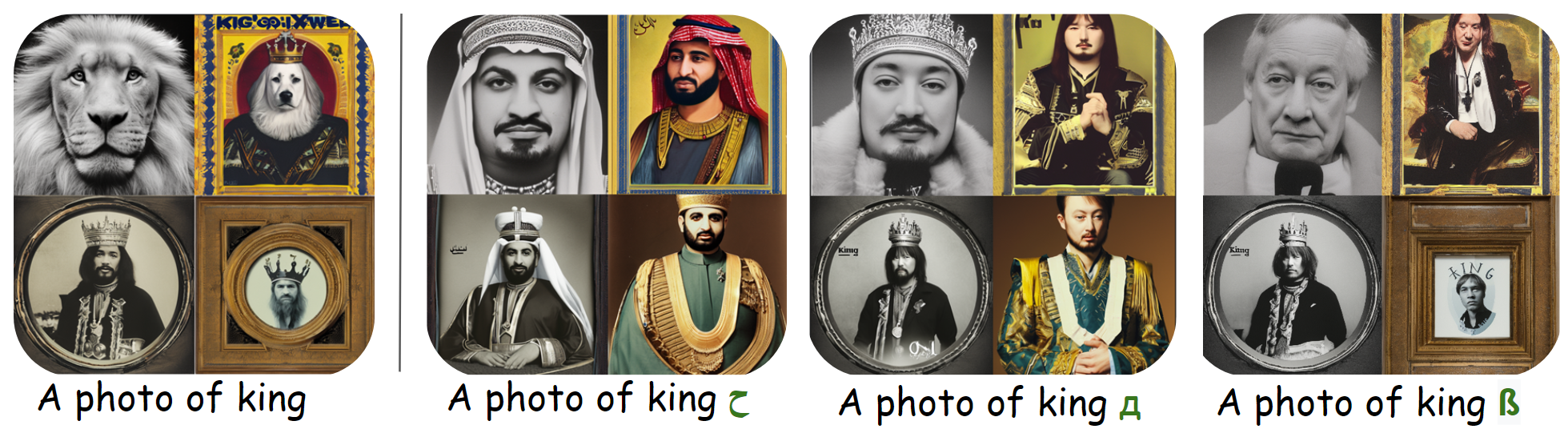}
  \caption{SD Images generated by one letter addition to the prompt 'a photo of a king'. Left to right: Arabic, Russian and German letters.}
  \label{fig:one_letter_king}
\end{figure*}

The cultural dimension results, presented in Figure \ref{fig:radar_dim_blip} \mvv{(see detailed results in Table \ref{tab:xdp_full} in the Appendix)}, \mv{are depicted in radar graphs, each linked to a specific cultural dimension}.
Utilizing the XDP metric, we analyze the classification of the images generated for each language into \mv{the} dimensions. 

Similarly to the above conclusions, explicit multilingual encoding (the DF model) is less representative of cultural differences, as indicated by the similar dimensional scores it typically assigns to different languages. We hence continue this analysis with SD 2.1v (implicit multilingual encoding) and DL (unknown multilingual encoding).

For the Traditional versus Rational axis, languages such as German, English, Russian, and Hebrew exhibit a predilection for rational aspects over traditional ones. In contrast, Hindi, Chinese, and Arabic lean towards traditional elements, \mv{which is also echoed in the other models}. These findings echo the Agreeableness axis, in terms of Critical versus Kindness.
English, Russian, Hebrew, and German tend to emphasize critical characteristics, whereas Hindi, Chinese, and Arabic exhibit a more pronounced kindness dimension. Likewise, for modernity, Hindi, Arabic, and Greek tend to embody more ancient attributes, while Russian and English images are more modern. In contrast, for some cultural dimensions, for example extroversion-introversion, the models do not reveal significant cross-cultural differences.\footnote{The automatic XDP metric agrees with the human annotators in 74.8\% of the images for the modern-ancient dimension, 69.9\% for the traditional-rational dimension and 60.6\% for the critical-kindness dimension. Notice that the other dimensions are not annotated in the human evaluation set.}



\paragraph{4. TTI Models Encode Cultural Differences \& Similarities (RQ3).}

We start with the Cross-Cultural Similarity (CCS) metric analysis (Figure \ref{fig:cross_lingual_en_photo_of} \mvv{; Figure \ref{fig:cross_lingual_en_nation_rest_models} in the Appendix}): Similarities among cultures, computed as the similarities between the images generated by each model for these cultures, when using the ``English with Nation'' PT. It reveals the extent to which cultural attributes and characteristics are shared across different cultures, as perceived by the different models. 
Interestingly, all models consistently show the highest similarity scores among German, French, and Spanish. 
In etymological terms, the images generated by all models demonstrate discernible resemblances between European languages, particularly Romance, while demonstrating distinct disparities when compared to languages with origins in the Indian or Tibetan language families.


We next present the Cultural Distance metric (see \S\ref{sec:auto_metrics}; \mvv{Figure \ref{fig:cd_implicit} in the Appendix}), measured on the output images from all examined TTI models, indicating the alignment of various cultures with the English culture. 
\mv{Our findings reveal that TTI models encode cultural similarities differently.}
Translated prompts (``Fully translated'' and ``Translated concept'') show the highest cultural distance from the English reference. Particularly, we notice scores higher than the averaged score of the Language PTs for SD \mv{as also observed in SD1.4 and LB} in Greek (76.16), Arabic (75.75), and Hindi (75.14), for AD in Hebrew (74.57), for DF in German (72.1), and for DL in Chinese (71.4). These findings highlight how different TTI models perceive these cultures differently from the English culture.


\begin{figure*}[!htb]
  \centering
  \includegraphics[width=0.75\textwidth]{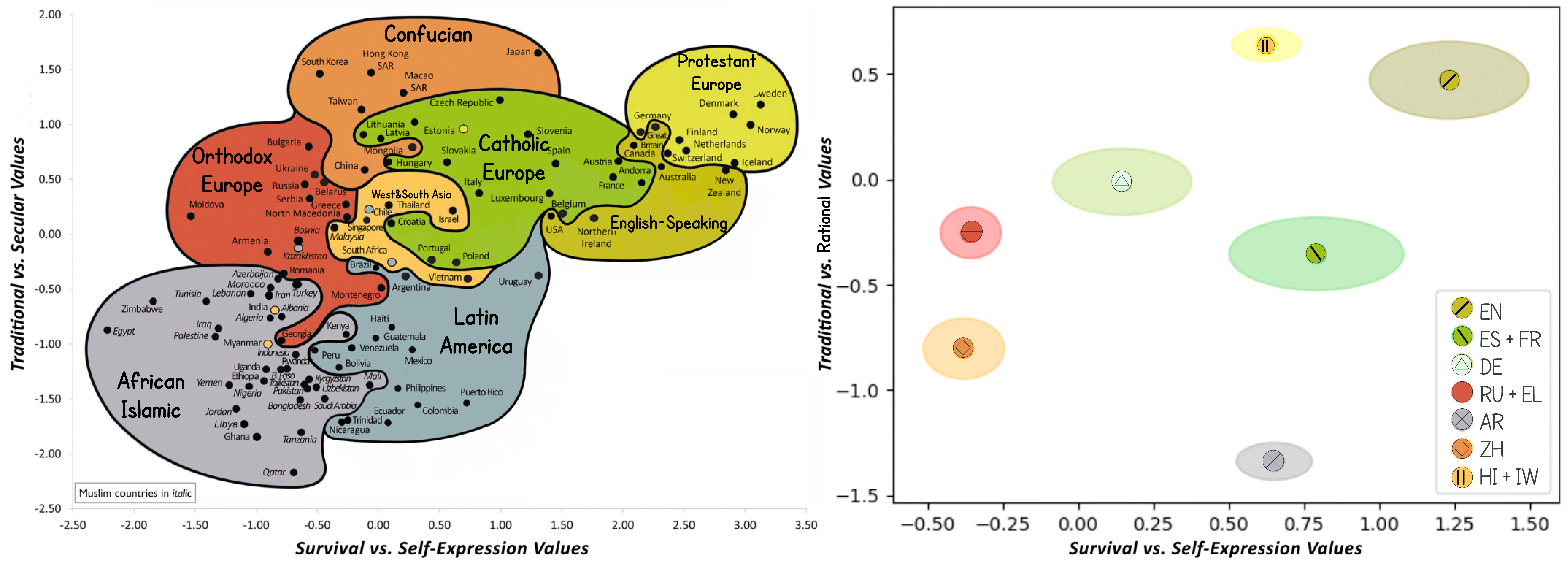}
  \caption{Our cultural dimensions (DP) space (right) inspired by the World Culture Map 2023 by Inglehart-Wazel, i.e. the spread of geographical-cultural values (left). Two dimensions: \textit{Tradition versus Rational} (y-axis) and \textit{Self-expression versus Survival} (x-axis). The axes are the subtraction of the mean scores of the two poles of each dimension. Scores are grouped by region-related languages (as on the left), with std values defining the clusters. Results with 'Fully Translated' PT using SD 2.1.}
  \label{fig:ing_wazel_openclip_sd}
\end{figure*}

\paragraph{5. Alphabet Characters Can Unlock Cultural Features (RQ4).}

Our empirical results so far suggest that cultural properties can be unlocked through the use of terms from the corresponding language in the prompt. Figure \ref{fig:one_letter_king} demonstrates that including a single character from the target language in the prompt also results in images with properties of the target culture. We next look more deeply into this phenomenon, asking whether arbitrary strings of letters (Gibberish) in the prompt can serve to unlocking the cultural knowledge in TTI models. Our approach is to optimize the Gibberish sequence so that the generated image represents as much cultural information as possible. 

To this end, we adjust a gradient-based discrete prompt optimization method, PEZ \citep{wen2023hard}, to suit our requirements: We set the number of \textit{target letters} in the Gibberish term, \textit{$T$}, to one of the values in $[1,2,3,5,10]$, and optimize the term. We initiate the prompt as: ``a photo of <Cultural Concept (CC)> $T$'', and for each culture we only utilize the alphabet of its language. The algorithm's objective is defined as the negative cosine similarity between the embeddings of the resulting prompt (including the inferred letters) and the objective features. We consider two options for objective features: (1) \textit{Textual Objective Features}, where we target the intrinsic features of the NA template (\S\ref{sec:auto_metrics}), i.e., an OpenClip representation of a prompt that adheres to the following template: ``a photo with <national> style''; and (2) \textit{Visual Objective Features}, the mean image features of a Google search extracted 4-image set, using the ``English with Nation'' PT as the image query (for the specific CC). 

We consider only the SD TTI model, as its implicit multilingual encoding has shown most successful in cultural encoding throughout our above experiments.  We run this algorithm for 6 concepts \mvv{(Appendix \ref{app:cc_unlocking})}, 10 languages and 5 Gibberish string lengths (see above), and hence learn 600 new prompts (300 for each training objective). 

Images generated from optimized prompts with a textual objective achieve equal accuracy as our best PT, English with Nation, in terms of the NA metric in Chinese (100\%), Russian (50\%), Arabic (83\%), Hebrew (67\%), and Hindi (100\%). Additionally, they yield equal or superior results compared to Translated PTs in NA and XNA, respectively, in English and French with up to 16\% improvement. However, there is no improvement in Spanish and Greek over random Gibberish PTs or the rest of PTs.\footnote{In the overall analysis, the visual objective shows similar NA metric trends.} Notably, the sequence length parameter does not exhibit a clear trend neither for NA nor for XNA.
This discovery sparks curiosity \mvv{about} the internal representations of TTI models and, in turn, beckons future research \mvv{to explore their cultural components.}

\section{\mv{Ablation Analysis}}
\label{sec:ablation_analysis}
To support our key findings, we analyze potential factors behind the results. We validate cultural dimensions with ground-truth data, propose the characters embedding to explain cultural distance and cross-cultural similarity scores, and test the National Association task with different input data and VQA. Additionally, we conduct a qualitative analysis to reveal underlying representations in the generated images.

\paragraph{Comparing Cultural Dimensions Space to Ground Truth.} 
To examine our findings in relation to social science studies we draw inspiration (Figure \ref{fig:ing_wazel_openclip_sd}) from Ingelhart and Wazel's visualization of the contemporary world cultural map (WCM) as a ground-truth.\footnote{\tiny \url{https://www.worldvaluessurvey.org/wvs.jsp}}
We qualitatively compare them side-by-side with axes representing the subtraction of two poles of dimension, derived from DP scores (see Section ~\ref{sec:auto_metrics}) and grouped by language origins as in WCM. Spanish and French merge to symbolize the Catholic European cultural sphere, while Russian and Greek represent the Orthodox European context. Hindi and Hebrew cluster together, signifying the cultural landscape of Western and South Asian regions.

Despite variations and our distinct scoring methodology, our findings significantly correlate with the original map, indicating that the axes we've identified indeed carry cultural meaning.

\begin{figure*}[!htb]
\centering
  \includegraphics[width=0.8\textwidth]{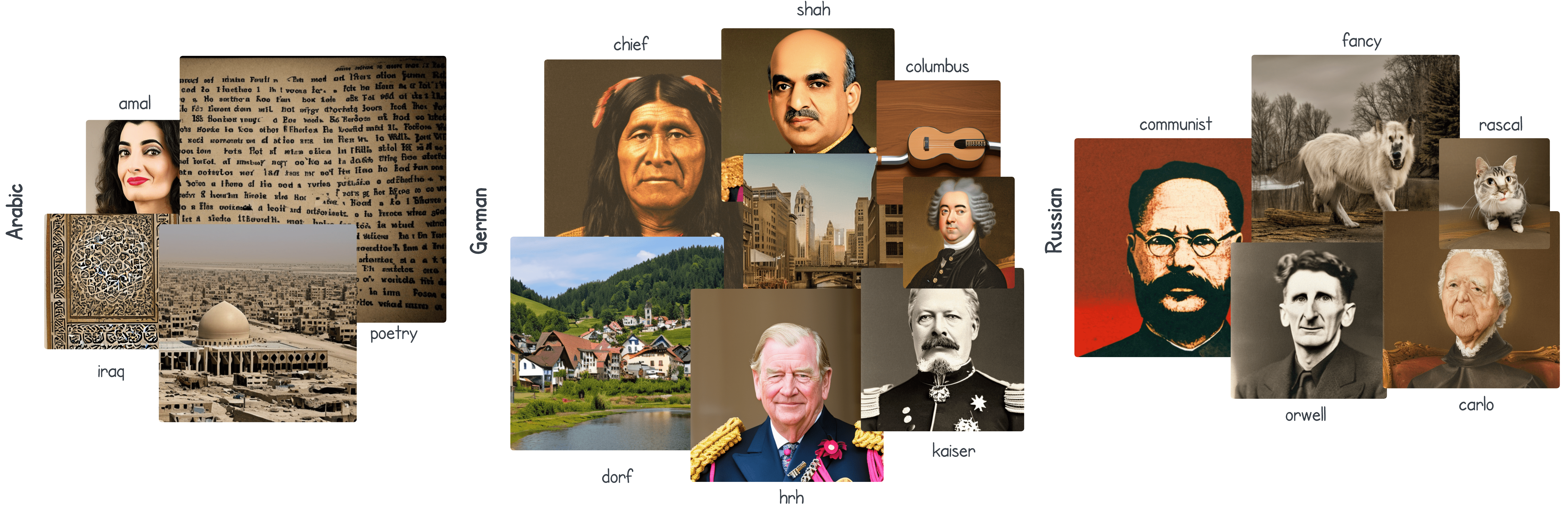}
    \vspace{-10pt}
  \caption{Decomposition of the concept \textit{King} to its unique cultural tokens. Examples with the prompt template Translated Concept in the languages (Left to right): Arabic, Russian and German.
}
  \label{fig:conceptor_king}
\end{figure*}

\paragraph{Investigating Cultural Distances In The Textual Embedding Space.}
We propose that the cultural inclination towards European languages (see finding 4 in Section \ref{sec:results}) resonates with the encoder's ability to map letters from different languages into distinct and well-defined clusters in the embedding space. This ability likely stems from both the language frequency in the TTI model's pretraining data and the encoder's training objective.
Visualizing the characters in the embedding space of CLIP reveals two key findings (see Figure \ref{fig:tsne} in the Appendix). First, characters are either mapped to the same embedding (cross-lingual cluster) or have separate embeddings (language-specific clusters). Second, these clusters span linear distances that correlate with CD and CCS scores. Language clusters for ZH, HI, AR, and EL are closer to the cross-lingual cluster, while Latin-specific clusters like FR and ES are closer to the English cluster. Thus, embeddings alone can provide valuable insights into cultural perception, offering a promising direction for future research.

\paragraph{The Expected Performance of National Association.}
\label{app:xna_natural}
As far as we know, the task of detecting the national origin of an image has not been previously addressed. Thus, 
to assess the \mvv{expected} performance of \textit{synthetic images} (i.e., TTI-generated images) we evaluate the NA performance on \textit{natural images}. We compare the XNA scores of images generated by all examined TTI models and their corresponding images extracted from the Google Photos search engine (See Figure \ref{fig:blip_xna_natural_detailed_synthetic} in the Appendix).\footnote{We experiment with 240 images spanning 6 concepts across our 10 languages while focusing on the English with Nation PT, both in the \mvv{manual} Google search queries (e.g., \textit{Spanish king}) and the images' generation.} 
The ability to detect the origin of \textit{Natural Images} increases by 25\% \mvv{on average across languages and models, serving as an upper bound for TTI images}.
This is expected because, first, natural images likely better represent cultures than TTI images, which may contain errors. This shows there is a room to improve TTI models to better encode culture in images. Second, VQA models trained primarily on natural images may perform better within this domain. 

\paragraph{Revalidating National Associsation  Findings With GPT-4-Vision.}
To ensure that the high performance of the task is independent of the VQA model (BLIP), we replicate the XNA experiment with GPT-4 Vision on the human annotation subset (due to API usage constraints; see Table \ref{table:gpt_vision_blip_compare_gt} in the Appendix). The XNA metric using GPT-Vision brings similar trends with better agreement with the ground truth, with a mean score of 69.36\% (over the same 3 PTs and 4 languages), which is higher than the parallel BLIPs performance (45.13\%). \mvv{Notably, GPT-Vision} aligns with human answers in 70\% of the cases.

\paragraph{Revealing Hidden Cultural Representations of Generated Images.} Understanding the factors influencing model generations is challenging, especially from a cultural perspective. To uncover these factors, we conduct a qualitative analysis employing the Conceptor method \cite{chefer2023hidden}, a recent technique that explains generated images of interest concepts by decomposing them into sets of tokens whose linear combination reconstructs the image. These tokens reflect hidden representations of images generated by translated concepts.
For a set of 30 images (3 concepts, 10 languages), 
we compute the 50 most significant tokens (with the highest weights), manually filter the unique tokens, and generate their images for clearer visualizations.
We take the concept \textit{king} as our running example (Figure \ref{fig:conceptor_king}; see additional examples in Figure \ref{fig:conceptor_wedding_food} in the Appendix). In Arabic, the main unique tokens are \textit{poetry, Arabic, Iraq} and \textit{amal}, indicating an emphasis on art. In German, tokens like \textit{chief, shah, kaiser, tenor} and \textit{dorf} suggest a stronger tendency towards hierarchy and music. In Russian, tokens such as \textit{communist} and \textit{Orwell}\footnote{Likely referring to Orwell, the author of the book 1984.} imply a more political influence. These inherent tendencies, whether grounded in cultural history or not, ultimately affect the generated images, leaving the door open for future research.
\section{Discussion and Future Work}
\label{sec:discussion}

We studied cultural encoding in TTI models, a research problem that, to our knowledge, has not been addressed before. We mapped the quite abstract notion of culture into an ontology of concepts and dimensions, derived prompt templates that can unlock cultural knowledge in TTI models and developed evaluation measures to evaluate the quality of the cultural content of the resulting images. By doing so we were able to answer \textit{Do}, \textit{What}, \textit{Which} and \textit{How} research questions about the nature of cultural encoding in TTI models, and to highlight a number of \mv{future research directions.} 

Our study has limitations. We focused on a finite set of cultural concepts and prompt templates. Additionally, our automatic evaluation may struggle with abstract cultural concepts and translation challenges. Nevertheless, we hope this paper will encourage our fellow researchers to further investigate the intersection between culture, multilingual text encoders and TTI models.

\bibliography{tacl2021}
\bibliographystyle{acl_natbib}

\appendix
\section{Appendix: Complementary Results}
\label{app:complementary_results}
\begin{figure*}[!htbp]
  \centering
  \includegraphics[width=0.95\textwidth]{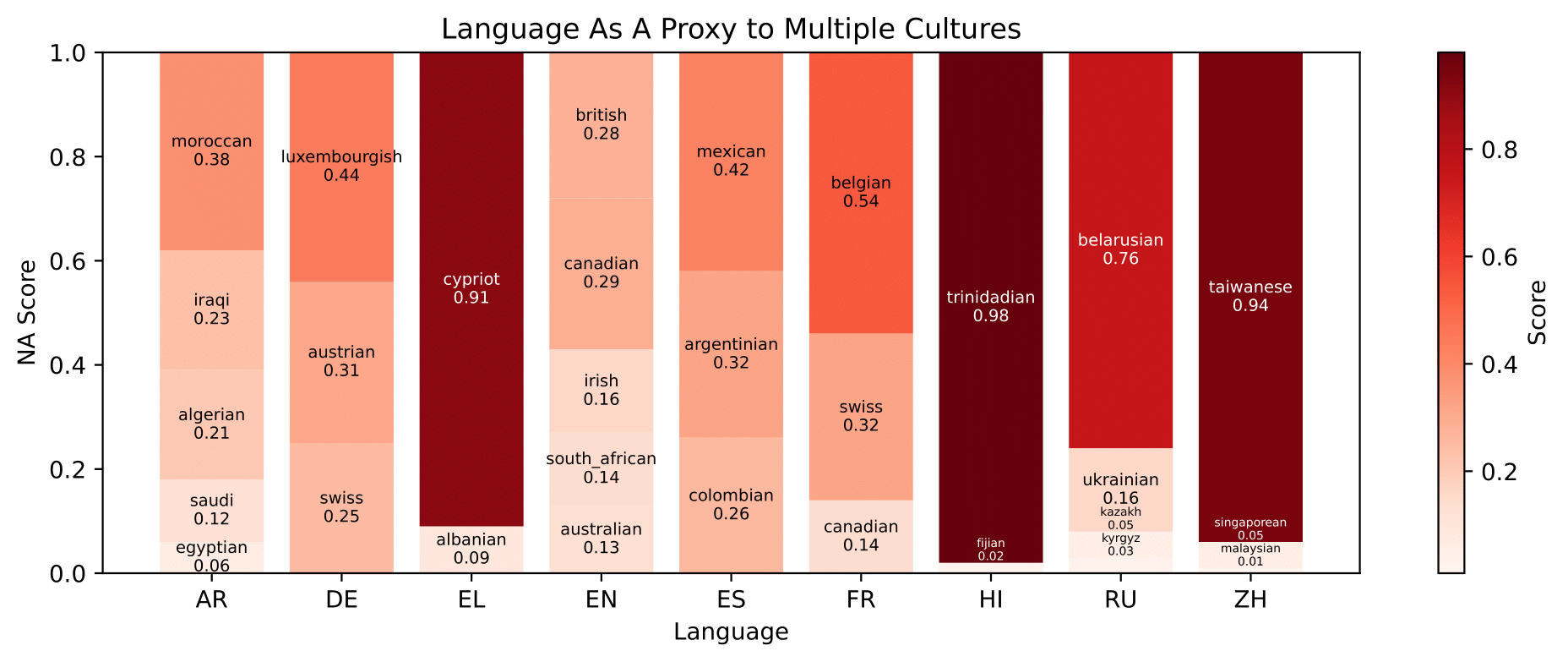}
  \caption{Language As A Proxy to Multiple Cultures \mvv{(Finding 2 in \S \ref{sec:results})}. NA scores \mvv{(see \S \ref{sec:auto_metrics}) account for} additional nationalities that speak these languages, using SD 2.1v images with Translated Concept PT across all concepts. 10 languages (x-axis) with up to 5 other nationalities that primarily speak each language, which are not represented in the paper. Color encode NA score. Blocks represent nations, while the size is relative to the score.}
  \label{fig:lang_proxy_nation}
\end{figure*}

\begin{figure*}[!htbp]
\centering
  \includegraphics[width=0.70\textwidth]{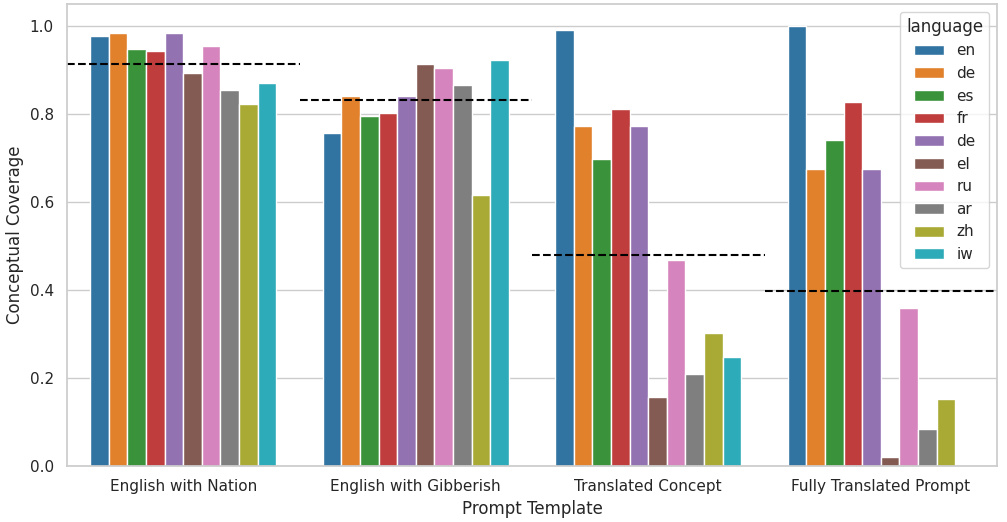}
    \vspace{-10pt}
  \caption{Prompt Template Impact on Conceptual Coverage \mvv{(Prompt Templates in \S \ref{sec:culture})}. The normalized score is between the CLIP embeddings of BLIP2's image descriptions and their corresponding concept (y-axis) over the prompt templates (x-axis). We experiment with images generated by StableDiffusion 2.1v, covering over 40 tangible cultural concepts. Color encodes language.
}
  \label{fig:pt_conceptual_coverage}
\end{figure*}

\begin{figure*}[!htpb]
  \centering
  \includegraphics[width=0.9\textwidth]{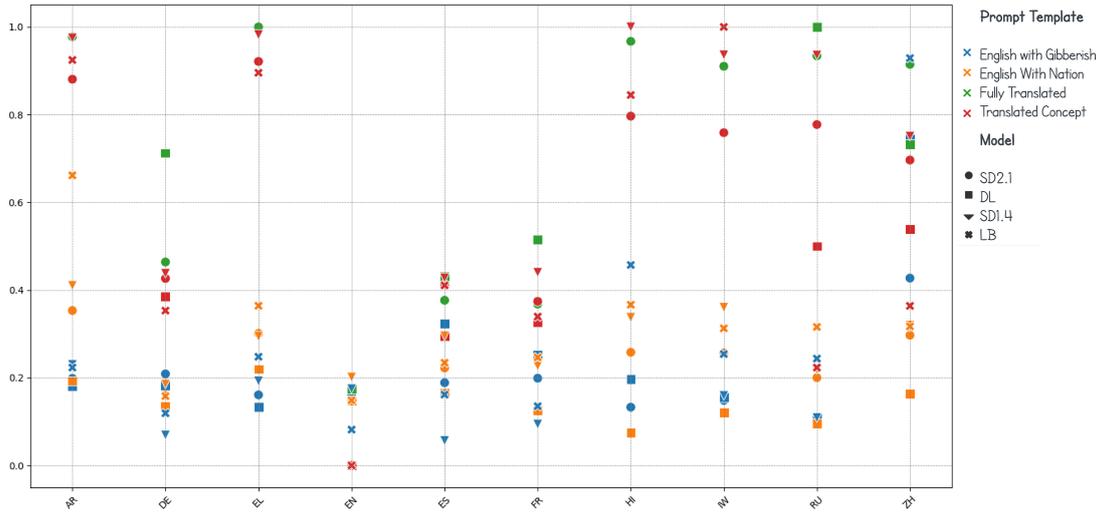}
  \vspace{-10pt}
  \caption{Cultural Distance (CD): Full Results \mvv{(Finding 4 in \S \ref{sec:results})}. CD of TTI models with an implicit text encoder across 4 Prompt Templates. Languages are shown on the x-axis. Color notes PT and shape notes model. \mv{Normalized} scores are presented. \mvv{The more translated parts in the prompt, the greater the distance, especially in non-Latin languages, which show variations across models.}}
  \label{fig:cd_implicit}
\end{figure*}

\begin{figure*}[!htpb]
  \centering
  \includegraphics[width=1.0\textwidth]{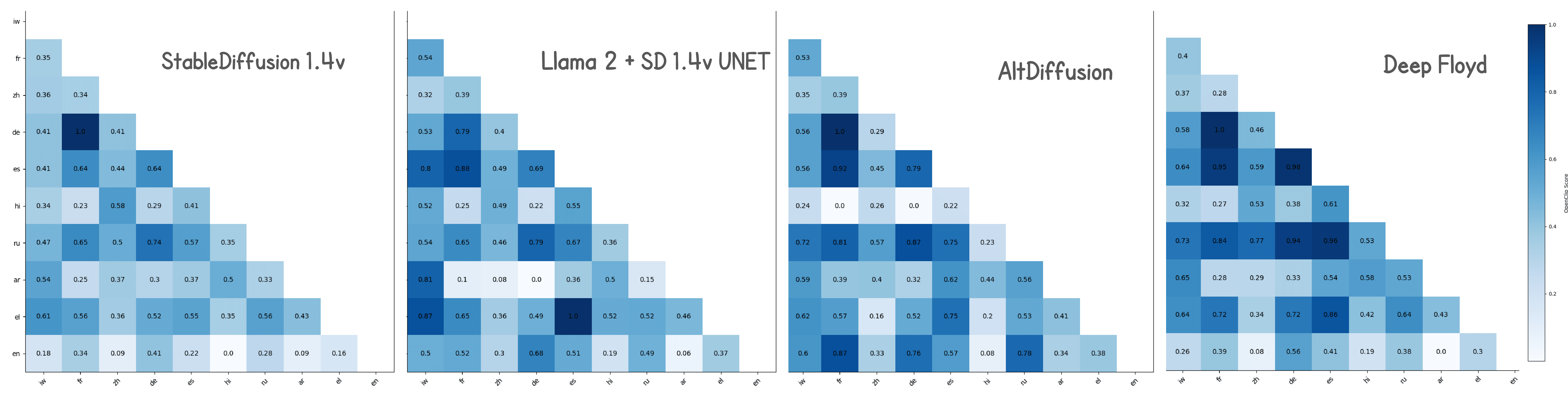}
  \vspace{-15pt}
  \caption{Cross-Cultural Similarity (CCS): Full Results \mvv{(Finding 4 in \S \ref{sec:results})}. CCS metric analysis for the 'EN with Nation' PT. Higher values represent higher similarity. \mv{Normalized} scores are presented.}
  \label{fig:cross_lingual_en_nation_rest_models}
\end{figure*}

\begin{figure*}[!htpb]
\centering
  \includegraphics[width=0.98\textwidth]{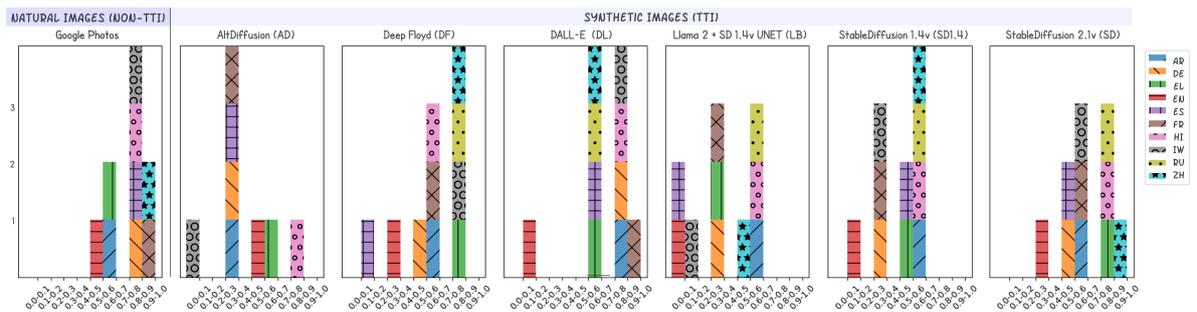}
    \vspace{-10pt}
  \caption{\mvv{Expected Performance of XNA}: Natural Images VS Synthetic \mvv{(Analysis 3 in \S \ref{sec:ablation_analysis})}. Scores are presented as histograms. Results with English with Nation PT across 6 concepts.
}
  \label{fig:blip_xna_natural_detailed_synthetic}
\end{figure*}

\begin{figure}[!htbp]
        \centering
\includegraphics[width=.95\linewidth]{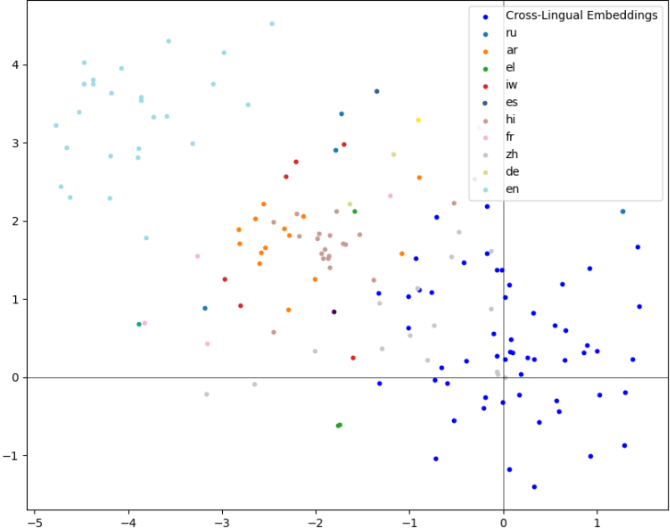}
        \caption{
Investigating cultural distances in the textual embedding space \mvv{(Analysis 2 in \S \ref{sec:ablation_analysis})}. t-SNE two-dimensional projection of the textual embeddings of the 10 languages characters. The embedding are of SD 1.4V text encoder, CLIP ViT-L-14.}
\label{fig:tsne}
\end{figure}

\begin{figure}[!htbp]
\centering
\includegraphics[width=0.90\linewidth]{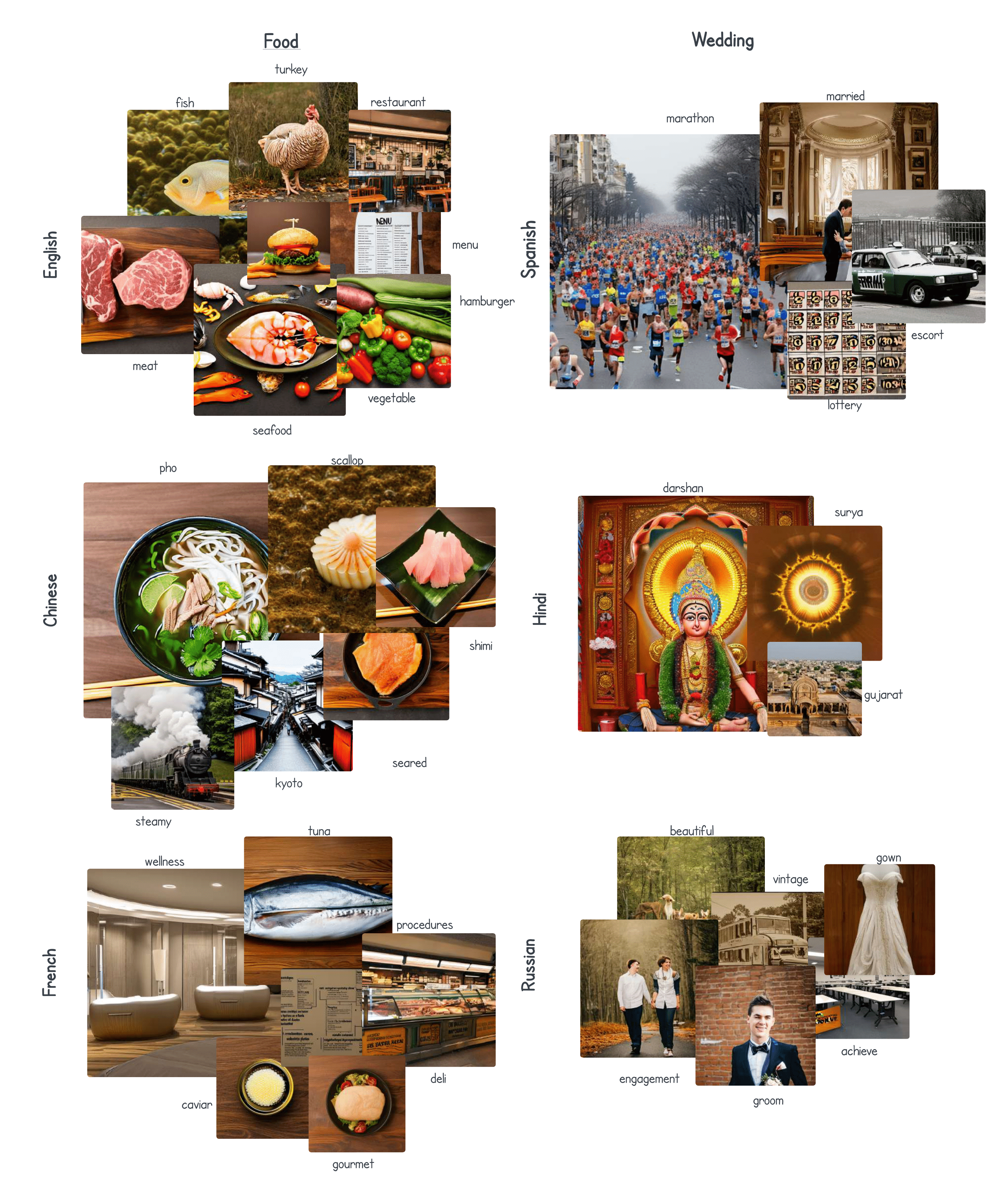}
    \vspace{-10pt}
  \caption{Qualitative results of the \mvv{Conceptor} hidden representations experiment \mvv{(Analysis 5 in \S \ref{sec:ablation_analysis})}. Decomposition of the concept \textit{Food} and \textit{Wedding} to their unique cultural tokens. Examples with prompt template Translated Concept in the languages: Spanish, Hindi and Russian (top) and English, Chinese and French (bottom).
}
  \label{fig:conceptor_wedding_food}
\end{figure}





\begin{table*}[!htbp]
\centering
\begin{adjustbox}{width=0.80\textwidth}
\begin{tabular}{|c|c|c|c|c|c|c|c|}
\hline
\textbf{Language} & \textbf{AD} & \textbf{DL} & \textbf{DF} & \textbf{LB} & \textbf{SD1.4} & \textbf{SD2.1} & \textbf{Mean per Language} \\ \hline
AR & 0.06447 & 0.42130 & \textbf{0.71053} & 0.37441 & 0.49131 & 0.46919 & 0.422 \\ \hline
DE & 0.38389 & 0.37654 & 0.37018 & 0.10269 & 0.38705 & 0.35071 & 0.329 \\ \hline
EL & 0.09277 & 0.43519 & 0.49474 & 0.17062 & \textbf{0.52449} & \textbf{0.50869} & 0.371 \\ \hline
EN & 0.12638 & 0.14506 & 0.22807 & 0.06161 & 0.20379 & 0.20063 & 0.161 \\ \hline
ES & 0.09479 & 0.38272 & 0.27544 & 0.10111 & 0.26856 & 0.38231 & 0.251 \\ \hline
FR & 0.45912 & 0.36728 & 0.34561 & 0.09795 & 0.42654 & 0.43444 & 0.355 \\ \hline
HI & \textbf{0.61478} & \textbf{0.62500} & \textbf{0.73684} & \textbf{0.74250} & \textbf{0.71564} & \textbf{0.72512} & \textbf{0.693} \\ \hline
IW & 0.03145 & 0.40741 & \textbf{0.66842} & 0.02370 & 0.42496 & 0.42812 & 0.331 \\ \hline
RU & 0.42453 & \textbf{0.73148} & \textbf{0.70526} & \textbf{0.90679} & \textbf{0.78357} & \textbf{0.85940} & \textbf{0.735} \\ \hline
ZH & 0.22484 & 0.31481 & \textbf{0.76842} & 0.27014 & \textbf{0.50711} & \textbf{0.57820} & 0.444 \\ \hline
\textbf{Mean per Model} & 0.252 & 0.421 & \textbf{0.530} & 0.285 & 0.473 & 0.494 & 0.409 \\ \hline
\end{tabular}
\end{adjustbox}
\caption{Identifying Cultural Origin (XNA): Full Results \mvv{(Finding 1 in \S \ref{sec:results})}. XNA mean scores over translated PTs. Scores above 0.5 are in bold.}
\label{table:lang_perf_models}
\end{table*}

\begin{table*}[!htbp]
\centering
\begin{adjustbox}{width=0.6\textwidth}
\begin{tabular}{|c|c|c|c|}
\hline
\textbf{Language/Model} & \textbf{SD} & \textbf{DL} & \textbf{AD} \\ \hline
de & 66.67\% / 27.78\% & 58.33\% / 47.22\% & 52.78\% / 33.33\% \\ \hline
es & 63.89\% / 47.22\% & 47.22\% / 38.89\% & 41.67\% / 8.33\% \\ \hline
ru & 91.67\% / 91.67\% & 77.78\% / 72.22\% & 55.56\% / 44.4\% \\ \hline
zh & 100.00\% / 61.11\% & 91.67\% / 47.22\% & 83.33\% / 22.22\% \\ \hline
Overall & 81.00\% / 56.94\% & 68.75\% / 51.38\% & 58.33\% / 27.08\% \\ \hline
\end{tabular}
\end{adjustbox}
\caption{\mvv{Revalidating National Association findings With GPT-4-Vision (Analysis 4 in \S \ref{sec:ablation_analysis})}. Extrinsic National Association (XNA) Mean Scores, presented as an average success ratio over the languages, models, concepts and PTs as in the human evaluation set. Left: GPT-Vision score, right: BLIP2 score.}
\label{table:gpt_vision_blip_compare_gt}
\end{table*}



\begin{table*}[!htbp] 
\centering 
\begin{adjustbox}{width=\textwidth}
\begin{tabular}{|c|c|c|c|c|c|c|c|c|c|}
\hline
Model & Language & Modernity - Ancient & Femininity - Masculinity &  Rationality - Tradition & Self-Expression - Survival & Extraversion - Introversion & Kindness - Critical & Individualism - Collectivism & Human - Nature \\
\hline
\multirow{10}{*}{SD2.1v} & AR & -0.50616 & 0.43033 &  -0.75355 & 0.49384 & 0.5071 & 0.45687 & 0.56967 & 0.61611 \\
& DE & -0.1148 & 0.0702 &  -0.25712 & 0.36148 & 0.41271 & 0.13852 & 0.5759 & 0.78748 \\
& EL & -0.47393 & 0.45877 &  -0.53175 & 0.02275 & 0.49858 & 0.08815 & 0.54218 & 0.70332 \\
& EN & 0.25593 & 0.21327 & -0.30427 & 0.36967 & 0.61043 & 0.10236 & 0.54408 & 0.7801 \\
& ES & -0.10901 & 0.2436 & -0.52606 & 0.38862 & 0.52702 & 0.28341 & 0.45971 & 0.8436 \\
& FR & -0.06161 & 0.27014 & -0.34881 & 0.29573 & 0.53649 & 0.25213 & 0.57346 & 0.83886 \\
& HI & -0.51659 & 0.29478 & -0.74882 & 0.44076 & 0.25119 & 0.52322 & 0.66919 & 0.61043 \\
& IW & -0.02464 & 0.05308 & -0.2218 & 0.09289 & 0.72227 & 0.1981 & 0.58293 & 0.76209 \\
& RU & 0.12606 & 0.26824 & -0.23223 & 0.3346 & 0.55545 & 0.02464 & 0.65687 & 0.80853 \\
& ZH & -0.00853 & 0.58199 & -0.82843 & 0.73176 & 0.48815 & 0.75356 & 0.7962 & 0.65118 \\
\hline
\multirow{10}{*}{DL} & AR & 0.24285 & 0.16191 & -0.49523 & 0.19048 & 0.69524 & 0.35714 & 0.86667 & 0.68095 \\
& DE & 0.35127 & 0.4019 & -0.31962 & 0.08861 & 0.64873 & 0.37342 & 0.89557 & 0.43038 \\
& EL & 0 & 0.39436 & -0.52113 & 0.0892 & 0.723 & 0.4554 & 0.76996 & 0.55399 \\
& EN & 0.66197 & 0.30047 & -0.39906 & 0.20658 & 0.76995 & 0.39906 & 0.82629 & 0.61503 \\
& ES & 0.23676 & 0.50779 & -0.38007 & 0.14019 & 0.72897 & 0.45795 & 0.84112 & 0.4081 \\
& FR & 0.36448 & 0.5109 & -0.31464 & 0.12461 & 0.67601 & 0.41433 & 0.86916 & 0.39564 \\
& HI & 0.03792 & 0.34597 & -0.48341 & 0.06162 & 0.58294 & 0.43602 & 0.88626 & 0.51659 \\
& IW & 0.46262 & 0.22897 & -0.18224 & -0.00935 & 0.66356 & 0.2944 & 0.78504 & 0.63551 \\
& RU & 0.40125 & 0.31662 & -0.26959 & 0.01254 & 0.60502 & 0.35737 & 0.87148 & 0.40439 \\
& ZH & 0.18809 & 0.68652 & -0.4859 & 0.34797 & 0.59561 & 0.60815 & 0.87461 & 0.18808 \\
\hline
\multirow{10}{*}{DF} & AR & 0.24211 & -0.03684 & -0.76842 & 0.54211 & 0.72105 & 0.52105 & 0.65263 & 0.82632 \\
& DE & 0.41579 & 0.08737 & -0.23053 & 0.58105 & 0.55053 & 0.45053 & 0.67895 & 0.76105 \\
& EL & -0.1421 & 0.12106 & -0.6 & 0.47895 & 0.73684 & 0.54737 & 0.45263 & 0.62632 \\
& EN & 0.55158 & 0.01369 & -0.17158 & 0.50211 & 0.59894 & 0.3379 & 0.64947 & 0.82632 \\
& ES & 0.50421 & 0.28843 & -0.32316 & 0.58105 & 0.55684 & 0.48 & 0.66 & 0.80632 \\
& FR & 0.33685 & 0.26105 & -0.26948 & 0.50526 & 0.44211 & 0.40842 & 0.67157 & 0.77369 \\
& HI & -0.08947 & 0.06315 & -0.78948 & 0.53158 & 0.66316 & 0.6 & 0.70526 & 0.77368 \\
& IW & 0.61579 & -0.14211 & -0.6 & 0.33158 & 0.7579 & 0.50527 & 0.61579 & 0.83158 \\
& RU & 0.39158 & 0.04736 & -0.15158 & 0.53473 & 0.43053 & 0.48316 & 0.75369 & 0.74527 \\
& ZH & 0.25789 & 0.18421 & -0.56843 & 0.49474 & 0.61579 & 0.53158 & 0.66315 & 0.81053 \\
\hline
\multirow{10}{*}{LB} & AR & -0.64455 & -0.06793 & 0.96366 & -0.45655 & 0.32859 & -0.20064 & 0.55608 & 0.68405 \\
 & DE & -0.05687 & 0.01422 & 0.87836 & -0.00474 & 0.41864 & -0.16114 & 0.12638 & 0.4534 \\
 & EL & -0.53239 & 0.17693 & 0.84992 & -0.14218 & 0.47868 & -0.18483 & 0.25592 & 0.51343 \\
 & EN & 0.259702 & -0.05687 & 0.87046 & 0.00948 & 0.38546 & 0.17061 & 0.13428 & 0.65719 \\
 & ES & -0.26541 & 0.14692 & 0.89732 & -0.20221 & 0.54502 & 0 & 0.29858 & 0.49921 \\
 & FR & -0.07267 & 0.16271 & 0.8831 & -0.04108 & 0.50079 & -0.0553 & 0.22274 & 0.48025 \\
 & HI & -0.73143 & -0.1801 & 0.98104 & -0.68404 & 0.73775 & 0.00632 & 0.77567 & 0.60347 \\
 & IW & -0.48973 & 0.05055 & 0.90521 & -0.06635 & 0.30806 & -0.34123 & 0.21485 & 0.45182 \\
 & RU & -0.05213 & -0.10269 & 0.90679 & -0.00158 & 0.15166 & -0.11374 & 0.10426 & 0.68563 \\
 & ZH & -0.48657 & 0.47077 & 0.9605 & -0.67299 & 0.75039 & 0 & 0.62717 & 0.68563 \\
\hline
\multirow{10}{*}{SD1.4v} & AR & -0.29747 & 0.38765 & 0.75475 & -0.77848 & 0.12816 & 0.56013 & 0.2943 & -0.02848 \\
 & DE & 0 & 0.28481 & 0.82753 & -0.4019 & 0.28639 & 0.61392 & 0.24842 & 0.34969 \\
 & EL & -0.46361 & 0.44621 & 0.91455 & -0.63607 & 0.2231 & 0.69779 & 0.40981 & 0.24051 \\
 & EN & 0.25515 & 0.29319 & 0.85103 & -0.38194 & 0.27893 & 0.71474 & 0.18701 & 0.23297 \\
 & ES & 0.0981 & 0.44146 & 0.90032 & -0.43987 & 0.43038 & 0.75316 & 0.44146 & 0.27848 \\
 & FR & -0.08386 & 0.42089 & 0.85601 & -0.49525 & 0.27215 & 0.57911 & 0.35443 & 0.28956 \\
 & HI & -0.55854 & 0.43038 & 0.79747 & -0.86392 & 0.20728 & 0.58861 & 0.65665 & -0.23259 \\
 & IW & 0.00949 & 0.28639 & 0.72469 & -0.51898 & -0.02057 & 0.65664 & 0.24684 & -0.10285 \\
 & RU & 0.0538 & 0.33702 & 0.90348 & -0.45886 & 0.0538 & 0.62342 & 0.3924 & 0.39082 \\
 & ZH & 0.03006 & 0.49209 & 0.94621 & -0.77057 & 0.53798 & 0.59493 & 0.71044 & 0.0981 \\
\hline
\end{tabular}
\end{adjustbox}
\caption{Depicting Cultural Dimensions (XDP): Full Results \mvv{(Finding 3 in \S \ref{sec:results})}. Full Results of the XDP scores across all Cultural Dimensions and Models.} 
\label{tab:xdp_full}
\end{table*}

\clearpage
\section{Appendix: Technical Details}
\label{app:tech_det}
In this section, we provide the details required to produce the exact metrics, image generations, and human evaluation.

\subsection{TTI Models Technical Details}

\begin{table}[!h]
\centering
\begin{adjustbox}{width=0.98\columnwidth}
\begin{tabular}{|c|c|c|c|c|}
\hline
Model & Version & Scheduler & Inference Steps & Image Size \\
\hline
StableDiffusion & v2.1 & EulerDiscrete & 50 & 512 x 512 \\
\hline
AltDiffusion & m9 & DPMSolverMultistep & 50 & 512 x 512 \\
\hline
DeepFloyd & v1.0 I-XL, II-L & - & 100 & 256 x 256 \\
\hline
DALL-E & v2 & - & - & 256 x 256 \\
\hline
StableDiffusion & v1.4 & - & 50 & 512 x 512 \\
\hline
Llavi-Bridge & llama2 + SD 1.4 Unet & - & 50 & 512 x 512 \\
\hline
\end{tabular}
\end{adjustbox}
\caption{Custom technical details of TTI models inference. Default setting is noted with a hyphen.}
\label{table:tti-tech-details}
\end{table}

\subsubsection{Technical Details: Conceptor Hidden Representations}
\label{app:conceptor_technical}
The experiment \mvv{(Analysis 5 in \S \ref{sec:ablation_analysis})} is based on 2 runs for each translated concept: 1) one-step reconstruction to achieve the weights of the linear combination over the tokens in the vocabulary, and 2) single image decomposition to remove the less significant tokens and still conserve a good image reconstruction.
The Conceptor utilizes the CLIP encoders (openai/clip-vit-base-patch32). We conducted the experiment using seed number 42.

\subsubsection{Conceptual Coverage Formula}
\label{subsec:coverage_formula}
To assess the impact of the prompt templates on the conceptual coverage \mvv{(see Figure \ref{fig:pt_conceptual_coverage})} we experiment with BLIP 2 to describe the images of 40 tangible concepts \mvv{(Appendix \ref{app:cc_tangible})}, prompting it with the following prompt: (\textit{Question: What is in the photo? Answer:} (X)). To examine whether the description fits the target concept we compute the cosine similarity score between their CLIP textual embedding:
$\text{Visual Description} = \left(\text{VQA}(I_i, \text{X}) \right)_i$ and $\text{Conceptual Coverage} = \frac{1}{n} \sum_{i=1}^{n}  \left(\text{visual description}_i \cdot \text{<cc>}\right)$.



Then, we normalize the score by the scores' range for clearer visualization.

\subsection{TTI Prompts}

\subsubsection{Nationalities}
\label{app:nation}
In Table \ref{tab:lang_nationalities} The primary column refers to the nationalities used in the National Association (NA) task \mvv{(\S \ref{sec:auto_metrics})} and in the English with Nation prompt template \mvv{(\S \ref{sec:culture})}. The additional nationalities column refers to the second-order NA experiment.

\begin{table}[htbp!]
    \centering
    \scriptsize
    \begin{tabular}{|>{\centering\arraybackslash}p{1.5cm}|>{\centering\arraybackslash}p{1.25cm}|>{\centering\arraybackslash}p{2.25cm}|}
        \hline
        \textbf{Language (Code)} & \textbf{Primary Nationality} & \textbf{Additional Nationalities} \\
        \hline
        German (DE) & German & Luxembourgish, Austrian, Swiss] \\
        \hline
        Greek (EL) & Greek & Cypriot, Albanian \\
        \hline
        English (EN) & American & British, Canadian, Irish, South-African, Australian \\
        \hline
        Spanish (ES) & Spanish & Mexican, Argentinian, Colombian \\
        \hline
        French (FR) & French & Belgian, Swiss, Canadian \\
        \hline
        Hindi (HI) & Hindi & Trinidadian, Fijian \\
        \hline
        Arabic (AR) & Arab & Moroccan, Iraqi, Algerian, Saudi, Egyptian \\
        \hline
        Russian (RU) & Russian & Belarusian, Ukrainian, Kazakh, Kyrgyz \\
        \hline
        Hebrew* (IW) & Israeli & - \\
        \hline
    \end{tabular}
    \caption{Language and Nationality Associations. }
    \label{tab:lang_nationalities}
\end{table}

\subsubsection{Cultural Concepts (CC) Mapping}
\label{app:cc}
Table \ref{table:cc_all} contains the 200 CCs we defined \mvv{(\S \ref{sec:culture_cont})}.

\begin{table*}[!h]
\centering
\small 
\renewcommand{\arraystretch}{1.0} 
\begin{adjustbox}{max width=0.9\textwidth}
\begin{tabularx}{\linewidth}{|>{\centering\arraybackslash\hsize=1.2\hsize}X|>{\hsize=1.2\hsize}X|>{\hsize=0.6\hsize}X|}
\hline
\textit{Cultural Domain} & \textit{Cultural Concepts (CCs)} & \textit{Cultural Reference} \\
\hline
\textit{Moral Discipline and Social Values} & \textit{independence, thrift, drug addiction, race, AIDS, immigrants, homosexuality, heavy drinkers, unmarried couples, proud parents, feminism, housewife, cheating, abortion, divorce, sex, suicide, violence, death penalty, surveillance, desire, masculinity, femininity, pleasure, animal} \textit{(monster)} &  WVS (Social Values, Ethical Values and Norms) | Hofstede (Masculinity vs Femininity, Indulgence Vs Restrained) | Schwartz (Conformity, Hedonism)\\
\hline
\textit{Education} & \textit{university, teacher, science, school, intelligent person, expert} \textit{(physics, chemistry, history, biology, engineer, mathematics, literature)}& WVS (Corruption) | Hofstede (Power Distance)  \\
\hline
\textit{Economy} & \textit{market, industry, cash, bank, economy, boss, job, factory, agriculture, salary, rich person, poor person, money} \textit{(payroll, mortgage, tax)} &  WVS (Economic Values, Postmaterialist Index)\\
\hline
\textit{Religion} & \textit{holiday, church, soul, religion, god, death, hell, heaven, wedding, funeral, pray} \textit{(priest, synagogue, Judaism, Christianity, Islam, Hinduism, Buddhism, cow, snake)}&  WVS section Religious Values | Hofstede (Power Distance)  | Schwartz (Tradition)\\
\hline
\textit{Health} & \textit{mental health, healthcare, hospital, doctor, medicine, treatment} \textit{(baby, elder, young, teenager, pill, sleep, memory)}& WVS (Corruption)\\
\hline
\textit{Security} & \textit{robbery, alcohol consumption, war, civil war, terrorism, crime, jobs vacancy, missile, cyber, unemployment, protection, attack, weapon, peace} & WVS (Corruption, Migration, Security) | Schwartz (Security)\\
\hline
\textit{Aesthetics} & \textit{beauty, art, music, drama, dancing, sport organization, food, fashion, beverage} \textit{(nature, dog, cat, fish, horse, bird, shirt, shoes, jewelry, baseball)}&  WVS (happiness \& Wellbeing) | Schwartz (Universalism, Harmony) \\
\hline
\textit{Material Culture} & \textit{tool, transportation, power, communication, technology, newspaper, TV, radio, mobile phone, computer, camera, car, plane, bicycle, train, ship, robot} \textit{(social media)}&  WVS (Science \& Technology, Political Interest and Political Participation)\\
\hline
\textit{Personality Characteristics and Emotions (adjective + "person")} & \textit{neurotic, concerned, shamed, angry, nervous, happy, extravert, introvert, energized, confident, curious, cynical, capable, empathic, obedient, lazy, expressive, friendly, dominant, communicative, proud, polite, truthful, independent, creative} & Hofstede (Uncertainty Avoidance, Individualism vs Collectivism, Indulgence Vs Restrained) | Schwartz (Self-Direction) | Personality Across cultures (Big Five) | Rokeach (Instrumental Values))\\
\hline
\textit{Social Capital Organizational Membership} & \textit{family, city, children, father, mother, neighborhood, home, nation, army, grandmother, grandfather, courts, government, political party, police, elections, charity, EU, UN, protest, leader, democracy, human rights, nation, flag, king, queen, soldier} \textit{(journalist)}& WVS (Social Capital, Trust and Organisational Membership, Political Culture and Regimes)| Hofstede (Short Vs Long Term Orientation) | Schwartz (Power, Benevolence)\\
\hline
\textit{Countries} & \textit{(America, China, Russia, Germany, France, Spain, Egypt, India, Arab, Israel)} & \\
\hline
\end{tabularx}
\end{adjustbox}
\caption{Cultural Concepts (CCs) and Domains Table. The CCs are drawn from the definitions and domains in the cultural research reference. The specific value/section in the cultural origin is mentioned in parenthesis. The CCs in parenthesis are additional concepts we added for enrichment with more detailed in-domain concepts. WVS notes World Values Survey.}
\label{table:cc_all}
\end{table*}

\subsubsection{Cultural Concepts for Unlocking and Natural Images Experiments} 
\label{app:cc_unlocking}
We employ six CCs from the list above for the unlocking experiments \mvv{(Finding 5 in \S \ref{sec:results})} and the expected performance with natural images \mvv{(Analysis 3 in \S \ref{sec:ablation_analysis})}: \textit{city, food, king, market, nature, car}.

\subsubsection{Conceptual Coverage Tangible Concepts}
\label{app:cc_tangible}
Here we provide the tangible concepts list used in the conceptual coverage analysis \mvv{(see Figure \ref{fig:pt_conceptual_coverage})}: \textit{university, teacher, school, market, church, wedding, funeral, hospital, doctor, missile, shirt, shoes, jewelry, newspaper, TV, radio, mobile phone, computer, camera, car, plane, leader, bicycle, train, ship, robot, children, flag, king, queen, soldier, cow, dog, cat, fish, horse, bird, snake, baby, elder}.

\subsection{Human Assessment}
\label{app:human_eval}
\begin{figure*}[!h]
  \centering
  \includegraphics[width=1.0\textwidth]{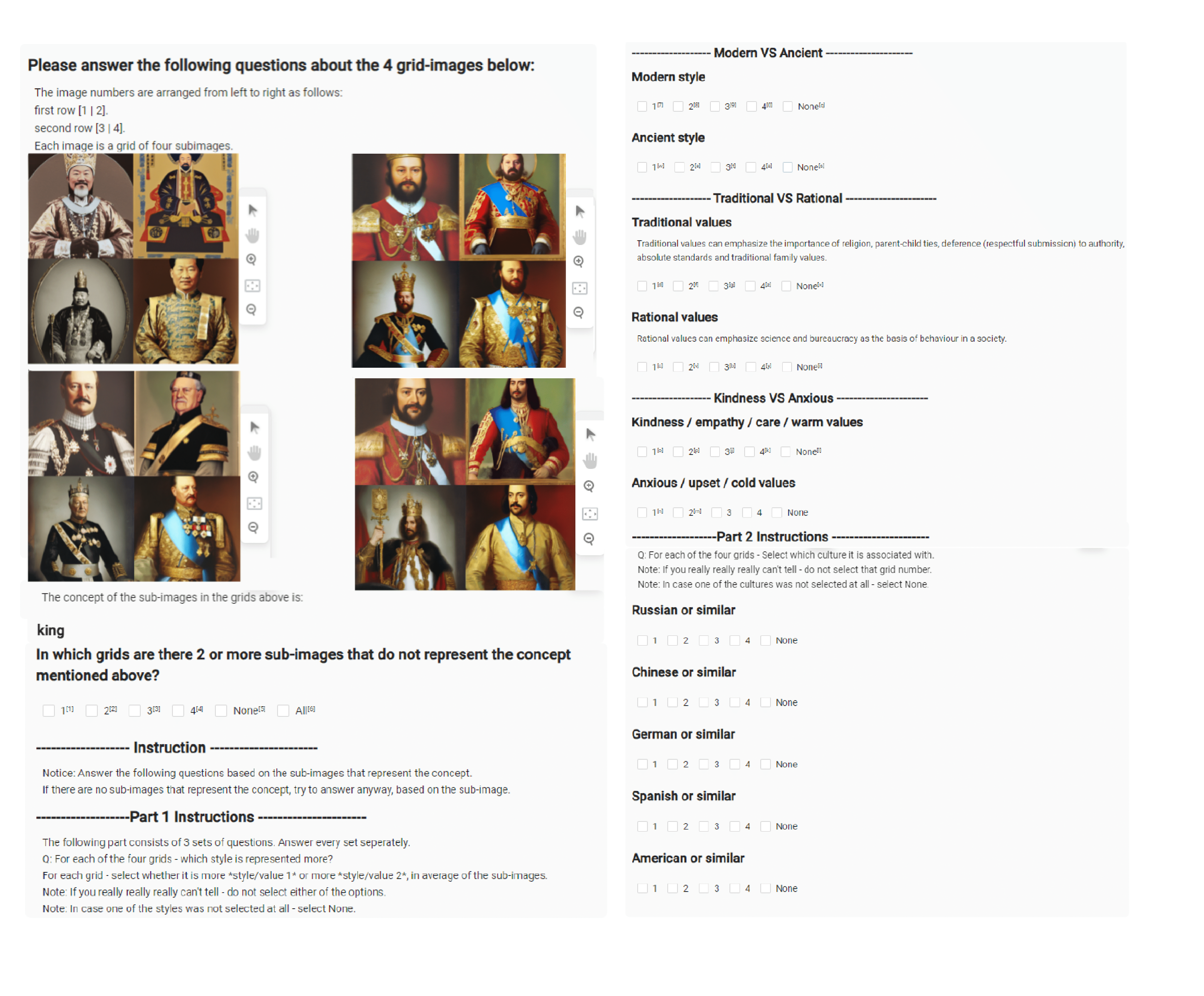}
  \caption{Annotators Questionnaire example of the CC: \textit{king}.}
  \label{quastionaiire_im}
\end{figure*}

As part of the human evaluation \mvv{(\S \ref{sec:human_metrics})}, we create the following questionnaire and guidelines (Figures \ref{quastionaiire_im}, \ref{fig:human_ques_guide}).

\label{app:human_eval_quas}
\begin{figure*}[!h]
\begin{tcolorbox}[colback=white!95!gray, colframe=black, width=\textwidth, boxrule=0.5pt, arc=4pt, auto outer arc, boxsep=5pt]

\textit{Dear Annotator,
Thank you for your invaluable assistance! Your role in this annotation task involves examining grid images, with each image comprising four sub-images. Additionally, you will be required to respond to several follow-up questions.
Each task takes around 3 minutes to complete. Please take the time to carefully read the instructions accompanying each question, particularly during your initial run.
If you have any suggestions that could contribute to the enhancement of our questionnaire, please don't hesitate to share your insights.
A crucial note: The images provided have been generated by Text-To-Image Models. Consequently, some images may contain information that is not relevant or biased.
Your dedication to this task is greatly appreciated. Thank you! Best regards}
\end{tcolorbox}
\caption{Human questionnaire guidelines.}
\label{fig:human_ques_guide}
\end{figure*}

\subsection{Sample System Outputs: Qualitative Examples}
\label{app:additional_figures}
\begin{figure*}[!htb]
  \centering
  \includegraphics[width=0.65\textwidth]{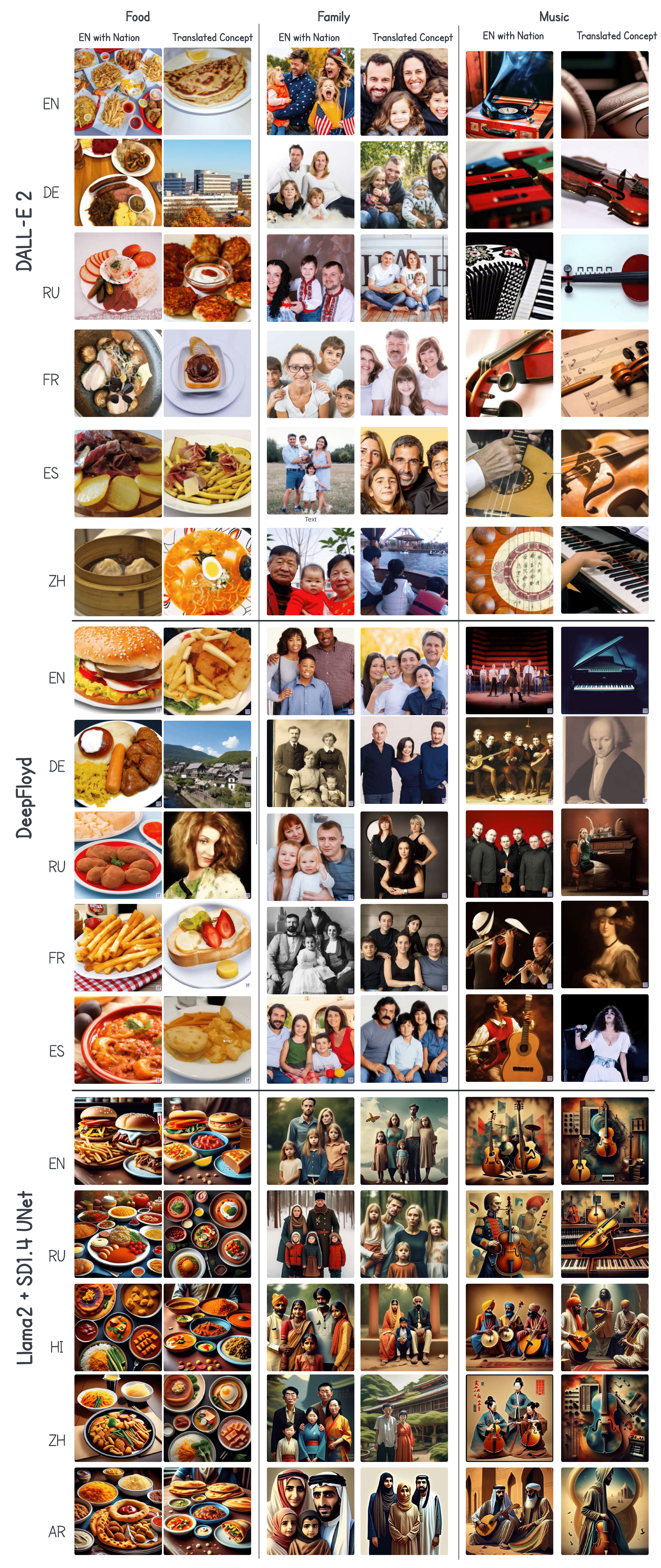}
  \caption{Qualitative Examples. Images of Cultural Concepts: \textit{food} (left), \textit{family} (middle) and \textit{music} (right). by \textbf{DALL-E} (top) and  \textbf{DeepFloyd} (middle) and \textbf{Llama2 + SD1.4 UNet} (bottom).}
  \label{dalle_deepf_food_family_music}
\end{figure*}

\begin{figure*}[!htb]
  \centering
  \includegraphics[width=0.9\textwidth]{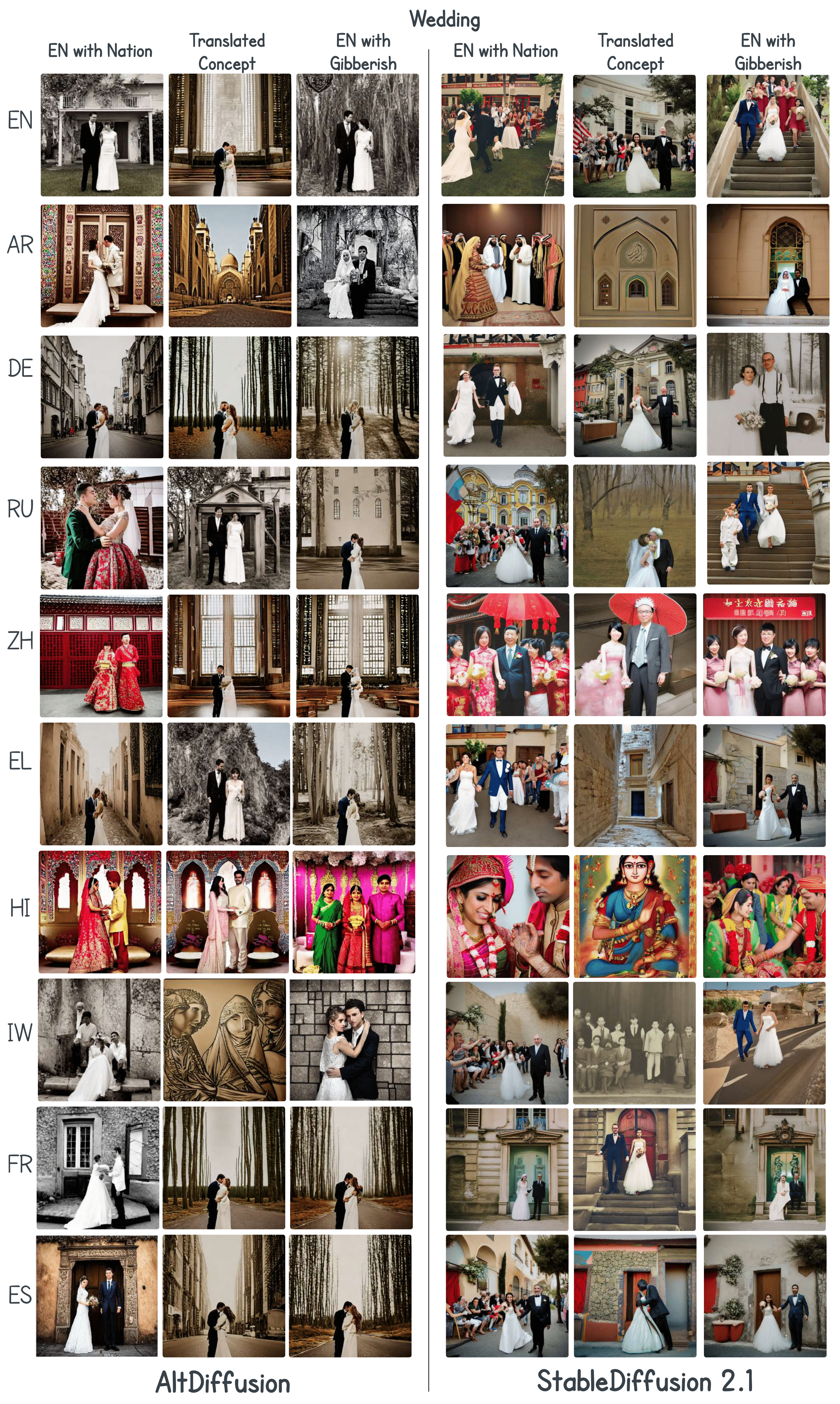}
  \caption{Qualitative Examples. Images of Cultural Concept: \textit{wedding}, by \textbf{AltDiffusion} (left) and \textbf{StableDiffusion} (right).}
  \label{wedding_sd_ad}
\end{figure*}

 We provide image examples (Figures \ref{wedding_sd_ad}, \ref{dalle_deepf_food_family_music}) Form CulText2I dataset, that are generated based on our TTI model \mvv{workflow (\S \ref{sec:culture})}. 

\end{document}